\documentclass[preprint,12pt]{elsarticle}

\newtheorem{theorem}{Theorem}
\newtheorem{lemma}{Lemma}
\newtheorem{definition}{Definition}
\newtheorem{corollary}{Corollary}

\usepackage{epsfig,amssymb,amsmath,amsfonts}

\journal{Information Sciences}

\begin{document}

\begin{frontmatter}



\title{Consistent transformations of belief functions}


\author{Fabio Cuzzolin}

\address{Department of Computing and Communication Technologies\\ Oxford Brookes University\\ Wheatley campus, Oxford OX33 1HX, United Kingdom}

\begin{abstract}
Consistent belief functions represent collections of coherent or non-contradictory pieces of evidence, but most of all they are the counterparts of consistent knowledge bases in belief calculus. The use of consistent transformations $cs[\cdot]$ in a reasoning process to guarantee coherence can therefore be desirable, and generalizes similar techniques in classical logics. Transformations can be obtained by minimizing an appropriate distance measure between the original belief function and the collection of consistent ones. We focus here on the case in which distances are measured using classical $L_p$ norms, in both the ``mass space" and the ``belief space" representation of belief functions. While mass consistent approximations reassign the mass not focussed on a chosen element of the frame either to the whole frame or to all supersets of the element on an equal basis, approximations in the belief space do distinguish these focal elements according to the ``focussed consistent transformation" principle. The different approximations are interpreted and compared, with the help of examples.
\end{abstract}


\end{frontmatter}


\section{Introduction}

Belief functions (b.f.s) \cite{Shafer76,dempster67multivariate} are complex objects, in which different and sometimes contradictory bodies of evidence may coexist, as they mathematically describe the fusion of possibly conflicting expert opinions or measurements. This is the case, for instance, of the pose estimation problem in computer vision \cite{cuzzolin14tfs}. There, image features do not necessary come from the object of interest, but may be extracted from the background: in such a case, the belief functions inferred from those feature will conflict \cite{cuzzolin13fusion}. Indeed, conflict and combinability play a central role in the theory of evidence \cite{yager87on,smets81degree,DBLP:journals/isci/RamerK93}, and have been recently subject to novel analyses \cite{1163941,hunter06fusion,lo06mss}.

\emph{Consistent knowledge bases and belief functions.} Making decisions based on objects collecting possibly inconsistent evidence, such as belief functions, is therefore less than trivial. This is a well known problem in classical logics, where the application of inference rules to inconsistent sets of assumptions or ``knowledge bases" may lead to incompatible conclusions, depending on the set of assumptions we start our reasoning from \cite{paris08unclog}. 
A variety of approaches have been proposed in this context to solve the problem of inconsistent knowledge bases, such as: fragmenting the latter into maximally consistent subsets; limiting the power of the formalism; or, adopting non-classical semantics \cite{priest89,batens00}. Even when a knowledge base is formally inconsistent, though, it may contain potentially useful information. Paris \cite{paris08unclog} tackles the problem by not assuming each proposition in the knowledge base as a fact, but by attributing to it a certain degree of belief in a probability logic approach. This leads to something similar to a belief function.

Indeed, extensions of classical logics in which propositions are assigned belief rather than truth values exist \cite{Saffiotti_abelief-function}. In this ``belief logic" framework, a belief function can be interpreted as the analogous of a knowledge base \cite{haenni05isipta}.\\
But then, what are the counterparts of consistent knowledge bases in the theory of evidence? To answer that we need to specify the notion of a belief function ``implying" a certain proposition. As we show here, under a rather sensible definition of such an implication, the class of belief functions which generalize consistent knowledge bases is uniquely determined as the set of b.f.s whose non-zero mass ``focal elements" have non-empty intersection. We are therefore allowed to call them \emph{consistent} belief functions (cs.b.f.s).

\emph{Consistent transformation.} 
Analogously to consistent knowledge bases, consistent b.f.s are characterized by null internal conflict. It may be therefore be desirable to transform a generic belief function to a consistent one prior to making a decision, or picking a course of action. A similar ``transformation" problem has been widely studied in both the probabilistic \cite{voorbraak89efficient,cobb03isf,daniel06on,cuzzolin07smcb} and possibilistic \cite{dubois90,dubois93possibilityprobability,baroni04ipmu} cases. A sensible approach, in particular, consists of studying the geometry \cite{black96examination,cuzzolin08smcc} of the class of b.f.s of interest and projecting the original belief function onto the corresponding geometric locus.\\
Indeed, consistent transformations can be built by solving a minimization problem of the form:
\begin{equation} \label{eq:cs-approx}
cs[b] = \arg\min_{cs \in \mathcal{CS}} dist(b,cs)
\end{equation}
where $b$ is the original belief function, $dist$ an appropriate distance measure between belief functions, and $\mathcal{CS}$ denotes the collection of all consistent b.f.s.

By plugging in different distance functions in (\ref{eq:cs-approx}) we get different consistent transformations. Indeed, Jousselme et al \cite{jousselme10belief} have recently conducted a very nice survey of the distance or similarity measures so far introduced in belief calculus, come out with an interesting classification, and proposed a number of generalizations of known measures. Many of these measures could be in principle employed to define indifferently consistent transformations or conditional belief functions \cite{cuzzolin10brest}, or approximate belief functions by necessity or probability measures. Other similarity measures between belief functions have been proposed by Shi et al \cite{shi10distance}, Jiang et al \cite{jiang08new}, and others \cite{khatibi10new,diaz06fusion}.

In this paper we focus on what happens when employing classical $L_p$ norms in the approximation problem. The $L_\infty$ norm, in particular, is closely related to consistent and consonant belief functions.\\ The region of consistent b.f.s can be expressed as:
\[
\mathcal{CS} = \Big \{ b : \max_{x \in\Theta} pl_b(x) = 1 \Big \},
\]
i.e., the set of b.f.s for which the $L_\infty$ norm of the ``plausibility distribution" or "contour function" $pl_b(x)$ is equal to 1. In addition, cs.b.f.s relate to possibility distributions, and possibility measures $Pos$ are inherently associated with $L_\infty$ as $Pos(A) = \max_{x \in A} Pos(x)$. In recent times, $L_p$ norms have been successfully employed in different problems such as probability \cite{cuzzolin07smcb} and possibility \cite{cuzzolin11isipta-consonant} transformation/approximation, or conditioning \cite{cuzzolin10brest,cuzzolin11isipta-conditional}.

As the author has proven \cite{cuzzolin08isaim-consistent}, geometrically, consistent (as well as consonant \cite{cuzzolin10fss}) belief functions live in a collection of simplices or ``simplicial complex" $\mathcal{CS}$. Each maximal simplex $\mathcal{CS}^{x}$ of the consistent complex is associated with associated with an ``ultrafilter" $\{ A \supseteq \{x\} \}$, $x \in \Theta$ of focal elements. A partial solution has therefore to be found separately for each maximal simplex of the consistent complex: these partial solutions are later compared to determine the global approximation(s).
In addition, geometric approximation can be performed in different Cartesian spaces. A belief function can be represented either by the vector $\vec{b} = [ b(A), \emptyset \subsetneq A \subsetneq \Theta ]'$ of its belief values, or the vector of its mass values $\vec{m}_b = [ m_b(A), \emptyset \subsetneq A \subsetneq \Theta ]'$. We call the set of vectors of the first kind \emph{belief space} $\mathcal{B}$ \cite{cuzzolin08smcc,cuzzolin01space}, and the collection of vectors of the second kind \emph{mass space} $\mathcal{M}$ \cite{cuzzolin10brest}.

\emph{Contribution and outline.} In this paper, we solve the $L_p$ consistent transformation problem in full generality in both the mass and the belief space, and discuss the semantics of the results.\\
After briefly recalling a few basis notions of the theory of evidence, we prove that consistent belief functions are the counterparts of consistent knowledge bases in belief logic (Section \ref{sec:consistent}). As we investigate the transformation problem in a geometric framework, we briefly recall in Section \ref{sec:approximation-complex} the geometry of the simplicial complex of consistent belief functions, and explain how we need to solve the approximation/transformation problem separately for each maximal simplex of this complex. We then proceed to solve the $L_1$-, $L_2$- and $L_\infty$-consistent approximation problems in full generality, in both the mass (Section \ref{sec:approximation-mass}) and the belief (Section \ref{sec:approximation-belief}) space representations. In Section \ref{sec:versus} we compare and interpret the outcomes of $L_p$ approximations in the two frameworks, with the help of the ternary example. We conclude and discuss the natural prosecution of this line of research in Section \ref{sec:conclusions}.

\section{Theory: consistent approximations of belief functions}

\subsection{Belief functions and belief logic} \label{sec:belief-functions}

A \emph{basic probability assignment} (b.p.a.) on a finite set or \emph{frame of discernment} $\Theta$ is a set function $m_b : 2^\Theta \rightarrow [0,1]$ on $2^\Theta \doteq \{ A \subseteq \Theta
\}$ s.t. $m_b(\emptyset) = 0$, $\sum_{A \subseteq \Theta} m_b(A) = 1$.
Subsets of $\Theta$ associated with non-zero values of $m_b$ are called \emph{focal elements}
(f.e.), and their intersection is called the \emph{core}:
\[
\mathcal{C}_b \doteq \bigcap_{A \subseteq \Theta : m_b(A) \neq 0} A.
\]
The \emph{belief function} (b.f.) $b:2^\Theta \rightarrow [0,1]$ associated with a basic probability
assignment $m_b$ on $\Theta$ is defined as: $\displaystyle b(A) = \sum_{B \subseteq A} m_b(B)$.\\
A dual mathematical representation of the evidence encoded by a belief function $b$ is the
\emph{plausibility function} (pl.f.) $pl_b : 2^{\Theta} \rightarrow [0,1]$, $A \mapsto pl_b(A)$
where the plausibility value $pl_b(A)$ of an event $A$ is given by $pl_b(A) \doteq 1 - b({A}^c) = \sum_{B \cap A \neq \emptyset} m_b(B)$ and expresses the amount of evidence \emph{not against} $A$.

Generalizations of classical logic in which propositions are assigned probability values have been proposed in the past. As belief functions naturally generalize probability measures, it is quite natural to define non-classical logic frameworks in which propositions are assigned \emph{belief values}, rather than probability values. This approach has been brought forward in particular by Saffiotti \cite{Saffiotti_abelief-function}, Haenni \cite{haenni05isipta}, and others.\\
In propositional logic, propositions or formulas are either true or false, i.e., their truth value $T$ is either 0 or 1 \cite{Mates72}. Formally, an \emph{interpretation} or \emph{model} of a formula $\phi$ is a valuation function mapping $\phi$ to the truth value ``true" (1). Each formula can therefore be associated with the set of interpretations or models under which its truth value is 1. If we define the frame of discernment of all the possible interpretations, each formula $\phi$ is associated with the subset $A(\phi)$ of this frame which collects all its interpretations.
If the available evidence allows to define a belief function on this frame of possible interpretations, to each formula $A(\phi) \subseteq \Theta$ is then naturally assigned a degree of belief $b(A(\phi))$ between 0 and 1 \cite{Saffiotti_abelief-function,haenni05isipta}, measuring the total amount of evidence supporting the proposition ``$\phi$ is true".

\subsection{Semantics of consistent belief functions} \label{sec:consistent}
In classical logic, a set $\Phi$ of formulas or ``knowledge base" is said to be \emph{consistent} if and only if there does not exist another formula $\phi$ such that the knowledge base implies both the formula and its negation: $\Phi \vdash \phi$, $\Phi \vdash \neg \phi$. Hence, it is impossible to derive incompatible conclusions from the set of propositions which form a consistent knowledge base. This is crucial if we want to derive univocal, non-contradictory conclusions from a given body of evidence.\\
A knowledge base in propositional logic $\Phi = \{ \phi : T(\phi) = 1 \}$ (where $T(\phi)$ denotes the truth value of proposition $\phi$) corresponds in belief logic \cite{Saffiotti_abelief-function} to a belief function, i.e., a set of propositions together with their non-zero belief values: $b = \{ A \subseteq\Theta : b(A) \neq 0 \}$. To determine what consistency amounts to in such a framework, we need to formalize the notion of \emph{proposition implied by a belief function}. One option is to decide that $b \vdash B\subseteq \Theta$ if $B$ is implied by all the propositions supported by $b$:
\begin{equation} \label{eq:implication-1}
b \vdash B \Leftrightarrow A \subseteq B \;\;\; \forall A : b(A) \neq 0.
\end{equation}
An alternative definition requires the proposition $B$ itself to receive non-zero support by the belief function $b$:
\begin{equation} \label{eq:implication-2}
b \vdash B \Leftrightarrow b(B)\neq 0.
\end{equation}
Whatever definition we choose for such implication relation, we can define the class of consistent belief functions as the set of b.f.s which cannot imply contradictory propositions.
\begin{definition} \label{def:consistent-1}
A belief function $b$ is consistent if there exists no proposition $A$ such that both $A$ and its negation $A^c$ are implied by $b$.
\end{definition}
When adopting the implication relation (\ref{eq:implication-1}), it is easy to see that $A \subseteq B$ $\forall A : b(A) \neq 0$ is equivalent to $\bigcap_{b(A)\neq 0} A \subseteq B$. Furthermore, as each proposition with non-zero belief value must by definition contain a focal element $C$ s.t. $m_b(C) \neq 0$, the intersection of all non-zero belief propositions reduces to that of all focal elements of $b$, i.e., the core of $b$:
\[
\bigcap_{b(A)\neq 0} A = \bigcap_{\exists C\subseteq A : m_b(C)\neq 0} A = \bigcap_{m_b(C)\neq 0} C = \mathcal{C}_b.
\]
Indeed, no matter our definition of implication, the class of consistent belief functions corresponds to the set of b.f.s whose core is not empty.
\begin{definition} \label{def:consistent-2}
A belief function is said \emph{consistent} if its core is non-empty.
\end{definition}
We can prove that, under either definition (\ref{eq:implication-1}) or definition (\ref{eq:implication-2}) of the implication $b \vdash B$, Definitions \ref{def:consistent-1} and \ref{def:consistent-2} are equivalent.
\begin{theorem} \label{the:equivalence}
A belief function $b : 2^\Theta \rightarrow [0,1]$ has non-empty core if and only if there do not exist two complementary propositions $A, A^c \subseteq \Theta$ which are both implied by $b$ in the sense (\ref{eq:implication-1}).
\end{theorem}
\emph{Proof.} A proposition $A$ is implied (\ref{eq:implication-1}) by $b$ iff $\mathcal{C}_b \subseteq A$. Accordingly, in order for both $A$ and $A^c$ to be implied by $b$ we would need $\mathcal{C}_b = \emptyset$. \hfill $\Box$

\begin{theorem} \label{the:incompatible}
A belief function $b : 2^\Theta \rightarrow [0,1]$ has non-empty core if and only if there do not exist two complementary propositions $A, A^c \subseteq \Theta$ which both enjoy non-zero support from $b$, $b(A) \neq 0$, $b(A^c) \neq 0$ (i.e., they are implied by $b$ in the sense (\ref{eq:implication-2})).
\end{theorem}
\emph{Proof.} By Definition \ref{def:consistent-1}, in order for a subset (or proposition, in a propositional logic interpretation) $A \subseteq \Theta$ to have non-zero belief value it has to contain the core of $b$: $A \supseteq \mathcal{C}_b$. In order to have both $b(A)\neq 0$, $b(A^c)\neq 0$ we need both to contain the core, but in that case $A \cap A^c \supseteq \mathcal{C}_b \neq \emptyset$ which is absurd as $A \cap A^c = \emptyset$.
\hfill $\Box$

\subsection{Achieving consistency in belief logic: consistent approximation}

Belief functions are complex objects, in which different and sometimes contradictory bodies of evidence coexist, as they may result from the fusion of possibly conflicting expert opinions and/or measurements. It is reasonable to conjecture that taking belief functions at face value may lead in some situations to incorrect/inappropriate decisions.\\
A variety of approaches have been proposed in the context of classical logics to solve the analogous problem with inconsistent knowledge bases: we mentioned some of them in the introduction \cite{priest89,batens00}. Paris' approach \cite{paris08unclog} is particularly interesting as it tackles the problem by attributing to each proposition in the knowledge base a certain degree of belief, leading to something similar to a belief function.\\ As consistent belief functions represent consistent knowledge bases in belief logic, such a mechanism can be be described there as an operator
\begin{equation} \label{eq:operator}
cs : \mathcal{B} \rightarrow \mathcal{CS}, \;\;\; b \mapsto cs[b]
\end{equation}
where $\mathcal{B}$ and $\mathcal{CS}$ denote respectively the set of all belief functions, and that of all cs.b.f.s. Consistent transformations (\ref{eq:operator}) can be built by posing a minimization problem of the form
\[
cs[b] = \arg\min_{cs \in \mathcal{CS}} dist(b,cs)
\]
where $b$ is the belief function to approximate and $dist$ is some distance measure between b.f.s. It is natural to pose this problem in a geometric setup.

\subsection{Geometry of belief functions: belief and mass vectors}

As belief functions $\displaystyle b:2^\Theta \rightarrow [0,1]$, $b(A) = \sum_{B\subseteq A} m_b(B)$ are set functions defined on a the power set $2^\Theta$ of a finite space $\Theta$, they are obviously completely defined by the associate set of $2^{|\Theta|} - 2$ belief values, that we can collect in a vector $\vec{b} = [ b(A), \emptyset \subsetneq A \subsetneq\Theta ]'$ (since $b(\emptyset) = 0$, $b(\Theta) = 1$ for all b.f.s). They can therefore be represented as points of $\mathbb{R}^{N-2}, N = 2^{|\Theta|}$ \cite{cuzzolin08smcc}. The set $\mathcal{B}$ of points of $\mathbb{R}^{N-2}$ which correspond to belief functions is an $N$-dimensional triangle or \emph{simplex} called \emph{belief space} \cite{cuzzolin08smcc}, namely:
$
\mathcal{B} = Cl(\vec{b}_A, \; \emptyset \subsetneq A \subseteq \Theta ),
$
where $Cl$ denotes the convex closure operator
\[
Cl(\vec{b}_1,...,\vec{b}_k) = \Big \{ \vec{b} \in \mathcal{B} : \vec{b} = \alpha_1 \vec{b}_1 + \cdots + \alpha_k \vec{b}_k, \sum_i \alpha_i = 1, \; \alpha_i\geq 0\; \forall i \Big \}
\]
and $\vec{b}_A$ is the vector associated with the categorical \cite{smets93belief} belief function $b_A$ assigning all the mass to a single subset $A\subseteq \Theta$: $m_{b_A}(A) = 1$, $m_{b_A}(B) = 0$ for all $B\neq A$. The vector $\vec{b} \in \mathcal{B}$ that corresponds to a belief function $b$ has coordinates $m_b(A)$ in the simplex $\mathcal{B}$ :
\[
\vec{b} = \sum_{\emptyset \subsetneq A \subseteq \Theta} m_b(A) \vec{b}_A.
\]
In the same way, each b.f. is uniquely associated with the related set of mass values $\{ m_b(A), \emptyset \subsetneq A \subseteq \Theta \}$ ($\Theta$ this time included). It can therefore be seen also as a point of $\mathbb{R}^{N-1}$, the vector $\vec{m}_b$ of its $N-1$ mass components:
\begin{equation} \label{eq:dev-m}
\vec{m}_b = \sum_{\emptyset \subsetneq A \subseteq \Theta} m_b(A) \vec{m}_A,
\end{equation}
where $\vec{m}_A$ is the vector of mass values associated with the categorical b.f. $b_A$. The collection $\mathcal{M}$ of points which are valid basic probability assignments is also a simplex, which we can call \emph{mass space}: $\mathcal{M} = Cl(\vec{m}_A, \emptyset \subsetneq A \subset \Theta)$.

\subsubsection{Binary example}

As an example let us consider a frame of discernment formed by just two elements, $\Theta_2 = \{x,y\}$. Each b.f. $b:2^{\Theta_2}\rightarrow [0,1]$ is completely determined by its mass/belief values $m_b(x)=b(x)$, $m_b(y)=b(y)$, as $m_b(\Theta) = 1 - m_b(x) - m_b(y)$ and $m_b(\emptyset)=0$. We can therefore collect them in a vector of $\mathbb{R}^{N-2} = \mathbb{R}^2$ (since $N = 2^2 = 4$): $\vec{m}_b = \vec{b} = [m_b(x), m_b(y)]' = \vec{b} \in \mathbb{R}^2$. In this example mass space and belief space coincide.
\begin{figure}[ht!]
\centering
\includegraphics[width = 0.65 \textwidth]{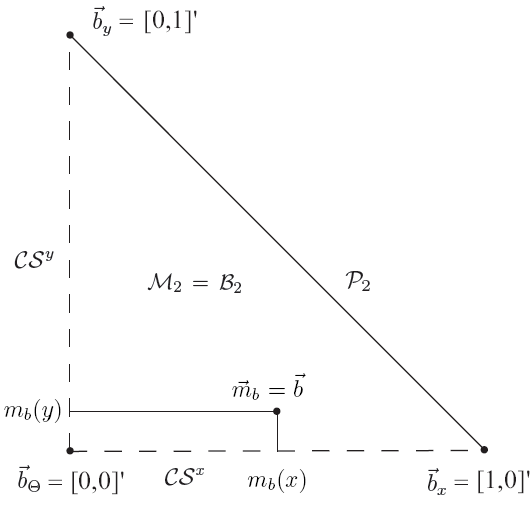}
\caption{\label{fig:b2} Both mass $\mathcal{M}_2$ and belief $\mathcal{B}_2$ spaces for a binary frame $\Theta = \{ x,y \}$ coincide with the triangle of $\mathbb{R}^2$ whose vertices are the mass (belief) vectors associated with the categorical b.f.s focused on $\{x\},\{y\}$ and $\Theta$, respectively. Consistent b.f.s live in the union of the two segments $\mathcal{CS}^x = Cl(\vec{m}_\Theta, \vec{m}_x) = Cl(\vec{b}_\Theta, \vec{b}_x)$ and $\mathcal{CS}^y = Cl(\vec{m}_\Theta, \vec{m}_y) = Cl(\vec{b}_\Theta, \vec{b}_y)$.}
\end{figure}

Since $m_b(x)\geq 0$, $m_b(y)\geq 0$, and $m_b(x) + m_b(y) \leq 1$ we can easily infer that the set $\mathcal{B}_2 = \mathcal{M}_2$ of all the possible basic probability assignments (belief functions) on $\Theta_2$ can be depicted as the triangle in the Cartesian plane of Figure \ref{fig:b2}, whose vertices are the points $\vec{b}_\Theta = \vec{m}_\Theta = [0,0]'$, $\vec{b}_x = \vec{m}_x = [1,0]'$, $\vec{b}_y = \vec{m}_y = [0,1]'$,
which correspond respectively to the vacuous belief function $b_\Theta$ ($m_{b_\Theta}(\Theta) = 1$), the Bayesian b.f. $b_x$ with $m_{b_x}(x) = 1$, and the Bayesian b.f. $b_y$ with $m_{b_y}(y) = 1$. The region $\mathcal{P}_2$ of all Bayesian b.f.s on $\Theta_2$ is the diagonal line segment $Cl(\vec{m}_x,\vec{m}_y) = Cl(\vec{b}_x,\vec{b}_y)$.\\
In the binary case consistent belief functions can have as list of focal elements either $\{\{x\},\Theta_2\}$ or $\{\{y\},\Theta_2\}$. Therefore the space of cs.b.f.s $\mathcal{CS}_2$ is the union of two line segments: $\mathcal{CS}_2 = \mathcal{CS}^x \cup \mathcal{CS}^y = Cl(\vec{m}_\Theta,\vec{m}_x) \cup Cl(\vec{m}_\Theta,\vec{m}_y) = Cl(\vec{b}_\Theta,\vec{b}_x) \cup Cl(\vec{b}_\Theta,\vec{b}_y)$.
It is easy to recognize that $\mathcal{CS}_2$ can be synthetically written in terms of the $L_\infty$ ($\max$) norm as
\begin{equation} \label{eq:cs2}
\mathcal{CS}_2 = \{ b : \min\{b(x),b(y)\} = 0 \} = \{ b : \max\{pl_b(x),pl_b(y)\} = 1 \}.
\end{equation}

\subsection{The consistent complex}

In the general case the geometry of consistent belief functions can be described by resorting to the notion of \emph{simplicial complex} \cite{Novikov_russian}. A simplicial complex is a collection $\Sigma$ of simplices of arbitrary dimensions possessing the following properties: 1. if a simplex belongs to $\Sigma$, then all its faces of any dimension belong to $\Sigma$; 2. the intersection of any two simplices is a face of both the intersecting simplices.\\ It has been proven that \cite{cuzzolin08isaim-consistent,cuzzolin09isipta-consistent} the region $\mathcal{CS}$ of consistent belief functions in the belief space is a simplicial complex, the union
\[
\mathcal{CS}_\mathcal{B} = \bigcup_{x \in \Theta} Cl(\vec{b}_A, A \ni x).
\]
of a number of (maximal) simplices, each associated with a ``maximal ultrafilter" $\{A \supseteq \{x\}\}$, $x \in\Theta$ of subsets of $\Theta$ (those containing a given element $x$). It is not difficult to see that the same holds in the mass space, where the consistent complex is the union
\[
\displaystyle \mathcal{CS}_\mathcal{M} = \bigcup_{x \in \Theta} Cl(\vec{m}_A, A \ni x)
\]
of maximal simplices $Cl(\vec{m}_A, A \ni x)$ formed by the mass vectors associated with belief functions whose core contains a particular element $x$ of $\Theta$.

\subsection{Using $L_p$ norms}

The geometry of the binary case hints to a close relationship between consistent belief functions and $L_p$ norms, in particular the $L_\infty$ one (Equation (\ref{eq:cs2})). It is easy to realize that this holds in general as, since the plausibility of all elements of their core is 1,
$
pl_b(x) = \sum_{A \supseteq \{x\}} m_b(A) = 1$ $\forall x \in \mathcal{C}_b,
$
the region of consistent b.f.s can be expressed as
$\displaystyle
\mathcal{CS} = \Big \{ b : \max_{x \in\Theta} pl_b(x) = 1 \Big \},
$
i.e., the set of b.f.s for which the $L_\infty$ norm of the "contour function" $pl_b(x)$ is equal to 1.\\ This argument is strengthened by the observation that cs.b.f.s relate to possibility distributions, and possibility measures $Pos$ are inherently related to $L_\infty$ as $Pos(A) = \max_{x \in A} Pos(x)$. It makes therefore sense to conjecture that a consistent transformation obtained by picking as distance function in the approximation problem (\ref{eq:cs-approx}) one of the classical $L_p$ norms maybe be meaningful. For vectors $\vec{m}_b$, $\vec{m}_{b'} \in \mathcal{M}$ representing the b.p.a.s of two belief functions $b$, $b'$, such norms read as:
\begin{equation} \label{eq:lp-m}
\begin{array}{lll}
\displaystyle \| \vec{m}_b - \vec{m}_{b'} \|_{L_1} & \doteq & \displaystyle \sum_{\emptyset \subsetneq B \subseteq \Theta} | m_b(B) - m_{b'}(B)|; \\ \| \vec{m}_b - \vec{m}_{b'} \|_{L_2} & \doteq & \displaystyle \sqrt{\sum_{\emptyset \subsetneq B \subseteq \Theta} ( m_b(B) - m_{b'}(B))^2}; \\ \displaystyle \| \vec{m}_b - \vec{m}_{b'} \|_{L_\infty} & \doteq & \displaystyle \max_{\emptyset \subsetneq B \subseteq \Theta} |m_b(B) - m_{b'}(B)|,
\end{array}
\end{equation}
while the same norms in the belief space read as:
\begin{equation} \label{eq:lp-b}
\begin{array}{c}
\displaystyle \| \vec{b} - \vec{b'} \|_{L_1} \doteq \sum_{\emptyset \subsetneq B \subseteq \Theta} | b(B) - b'(B)|; \; \| \vec{b} - \vec{b'} \|_{L_2} \doteq \sqrt{\sum_{\emptyset \subsetneq B \subseteq \Theta} ( b(B) - b'(B))^2}; \\ \displaystyle \| \vec{b} - \vec{b'} \|_{L_\infty} \doteq \max_{\emptyset \subsetneq B \subseteq \Theta} |b(B) - b'(B)|.
\end{array}
\end{equation}

In recent times, $L_p$ norms have been employed in different problems such as probability \cite{cuzzolin07smcb} and possibility \cite{cuzzolin11isipta-consonant} transformation/approximation, or conditioning \cite{cuzzolin10brest,cuzzolin11isipta-conditional}. In the probability transformation problem \cite{voorbraak89efficient,cobb03isf,smets88beliefversus}, $p[b] = \arg\min_{p \in \mathcal{P}} dist(b,p)$, the use of $L_p$ norms leads indeed to quite interesting results. The $L_2$ approximation produces the so-called ``orthogonal projection" of $b$ onto $\mathcal{P}$ \cite{cuzzolin07smcb}. In addition, the set of $L_1$/$L_\infty$ probabilistic approximations of $b$ coincide with the set of probabilities consistent with $b$: $\{ p : p(A) \geq b(A) \}$ (at least in the binary case).\\ Consonant approximations of belief functions obtained by minimizing $L_p$ distances in the mass space have simple interpretations in terms of redistribution of the mass outside a desired chain $A_1 \subset \cdots \subset A_n$, $|A_i|=i$ of focal elements to a single element of the chain, or all in equal terms \cite{cuzzolin11isipta-consonant}. Conditional b.f.s  can also be defined by minimizing $L_p$ distances from a ``conditioning simplex"
$
\mathcal{B}_A = Cl(\vec{b}_B, \emptyset\subsetneq B \subseteq A) \big / \mathcal{M}_A = Cl(\vec{m}_B, \emptyset\subsetneq B \subseteq A)
$
in the belief/mass space determined by the conditioning event $A$ \cite{cuzzolin10brest}. In the mass space, the obtained conditional b.f.s have natural interpretation in terms of Lewis' ``general imaging" \cite{lewis76,Gardenfors} applied to belief functions.
\begin{figure}[ht!]
\centering
\begin{tabular}{c}
\includegraphics[width = 0.75 \textwidth]{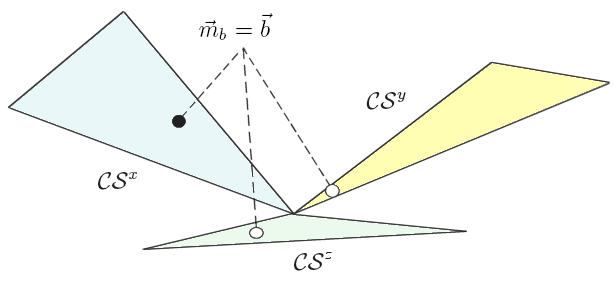} 
\end{tabular}
\caption{\label{fig:approx-complex} Left: To minimize the distance of a point from a simplicial complex, we need to find all the partial solutions (\ref{eq:cs-approx-partial}) on all the maximal simplices in the complex (empty circles), to later compare these partial solutions and select a global optimum (black circle).}
\end{figure}

\subsection{Consistent approximation on the complex} \label{sec:approximation-complex}

As the consistent complex $\mathcal{CS}$ is a \emph{collection} of linear spaces (better, simplices which generate a linear space), solving the problem (\ref{eq:cs-approx}) involves finding a number of partial solutions in the belief/mass space
\begin{equation} \label{eq:cs-approx-partial}
cs^x_{\mathcal{B},L_p}[b] = \arg\min_{\vec{cs} \in \mathcal{CS}_\mathcal{B}^x} \| \vec{b} - \vec{cs} \|_{L_p}, \;\;\; cs^x_{\mathcal{M},L_p}[m_b] = \arg\min_{\vec{m}_{cs} \in \mathcal{CS}_\mathcal{M}^x} \| \vec{m} - \vec{m}_{cs} \|_{L_p},
\end{equation}
respectively (see Figure \ref{fig:approx-complex}-left). Then, the distance of $b$ from all such partial solutions has to be assessed in order to select a global approximation.

\section{Calculation: consistent approximation in $\mathcal{M}$} \label{sec:approximation-mass}

Let us compute the analytical form of all $L_p$ consistent approximations in the mass space. We start by describing the difference vector $\vec{m}_b - \vec{m}_{cs}$ between the original mass vector and its approximation. Using the notation
\[
\vec{m}_{cs} = \sum_{B \supseteq \{x\}, B \neq \Theta} m_{cs}(B) \vec{m}_B, \;\;\; \vec{m}_b = \sum_{B \subsetneq \Theta} m_b(B) \vec{m}_B
\]
(as in $\mathbb{R}^{N-2}$ $m_b(\Theta)$ is not included by normalization) the difference vector is
\begin{equation} \label{eq:difference-vector-m}
\vec{m}_b - \vec{m}_{cs} = \sum_{B \supseteq \{x\}, B \neq \Theta} (m_b(B) - m_{cs}(B)) \vec{m}_B + \sum_{B \not \supset \{x\}} m_b(B) \vec{m}_B
\end{equation}
so that its classical $L_p$ norms read as
\begin{equation} \label{eq:norms-in-M}
\begin{array}{lll}
\| \vec{m}_b - \vec{m}_{cs} \|^{\mathcal{M}}_{L_1} & = & \displaystyle \sum_{B \supseteq \{x\}, B \neq \Theta } |m_b(B) - m_{cs}(B)| + \sum_{B \not \supset \{x\}} | m_b(B) |,  \\ \| \vec{m}_b - \vec{m}_{cs} \|^{\mathcal{M}}_{L_2} & = & \displaystyle \sqrt{ \sum_{B \supseteq \{x\}, B \neq \Theta } |m_b(B) - m_{cs}(B)|^2 + \sum_{B \not \supset \{x\}} | m_b(B) |^2 }, \\ \| \vec{m}_b - \vec{m}_{cs} \|^{\mathcal{M}}_{L_\infty} & = & \displaystyle \max \Big \{ \max_{B \supseteq \{x\}, B \neq \Theta } |m_b(B) - m_{cs}(B)|, \max_{B \not \supset \{x\}} | m_b(B) | \Big \}.
\end{array}
\end{equation}

\subsection{$L_1$ approximation} \label{sec:l1-m}

Let us tackle first the $L_1$ case. After introducing the variables $\beta(B) \doteq m_b(B) - m_{cs}(B)$ we can write the $L_1$ norm of the difference vector as
\begin{equation} \label{eq:l1norm}
\| \vec{m}_b - \vec{m}_{cs} \|^{\mathcal{M}}_{L_1} = \sum_{B \supseteq \{x\}, B \neq \Theta } |\beta(B)| + \sum_{B \not \supset \{x\}} | m_b(B) |,
\end{equation}
which is obviously minimized by $\beta(B) = 0$, for all $B \supseteq \{x\}$, $B \neq \Theta$. Hence:
\begin{theorem} \label{the:l1-m}
Given an arbitrary belief function $b:2^\Theta \rightarrow [0,1]$ and an element $x \in \Theta$ of the frame, its unique partial $L_1$ consonant approximation $cs^x_{\mathcal{M},L_1}[m_b]$ in $\mathcal{M}$ with core containing $x$ is the consonant b.f. whose mass distribution coincides with that of $b$ on all the subsets containing $x$:
\begin{equation}\label{eq:l1-m}
m_{cs^x_{\mathcal{M},L_1}[m_b]}(B) = \left \{ \begin{array}{ll} m_{b}(B) & \forall B \supseteq \{x\}, B \neq \Theta \\ m_b(\Theta) + b(\{x\}^c) & B=\Theta. \end{array} \right .
\end{equation}
\end{theorem}
The mass value for $B = \Theta$ comes from normalization, as follows:
\[
m_{cs^x_{\mathcal{M},L_1}[m_b]}(\Theta) = 1 - \sum_{B \supseteq \{x\}, B \neq \Theta} m_{cs^x_{\mathcal{M},L_1}[m_b]}(B) =  m_b(\Theta) + b(\{x\}^c).
\]
The mass of all the subsets not in the desired ``principal ultrafilter" $\{ B \supseteq \{x\} \}$ is simply re-assigned to $\Theta$. A similarity emerges with the case of $L_1$ conditional belief functions \cite{cuzzolin10brest}, when we recall that the set of $L_1$ conditional belief functions $b_{L_1,\mathcal{M}}(.|A)$ with respect to $A$ in $\mathcal{M}$ is the simplex whose vertices are each associated with a subset $\emptyset \subsetneq B \subseteq A$ of the conditional event $A$, and have b.p.a. (compare Equation (\ref{eq:l1-m})):
\[
\left \{ \begin{array}{ll} m'(B) = m_b(B) + 1 - b(A), & \\ m'(X) = m_b(X) & \forall \emptyset \subsetneq X \subsetneq A, X\neq B. \end{array} \right.
\]
In the $L_1$ conditional case, each vertex of the set of solutions is obtained by re-assigning the mass \emph{not in the conditional event $A$} to a single subset of $A$, just as in $L_1$ consistent approximation all the mass \emph{not in the principal ultrafilter} $\{ B \supseteq \{ x \} \}$ is re-assigned to the top of the ultrafilter, $\Theta$.

\subsubsection{Global approximation} The global $L_1$ consistent approximation in $\mathcal{M}$ coincides with the partial approximation (\ref{eq:l1-m}) at minimal distance from the original mass vector $\vec{m}_b$. By (\ref{eq:l1norm}) the partial approximation focussed on $x$ has distance $b(\{x\}^c) = \sum_{B \not \supset \{x\}} m_b(B)$ from $\vec{m}_b$.\\ The global $L_1$ approximation(s) form therefore the union of the partial approximation(s) associated with the maximal plausibility singleton(s):
\begin{equation} \label{eq:global-l1-m}
cs_{L_1,\mathcal{M}}[m_b] = \bigcup_{\arg \min_{x} b(x^c) = \arg \max_x pl_b(x)} cs^x_{\mathcal{M},L_1}[m_b].
\end{equation}
This is in accordance with our intuition, as it makes sense to focus on the singletons which are supported by the strongest evidence.

\subsubsection{A running example}

\begin{figure}[ht!]
\centering
\begin{tabular}{c}
\includegraphics[width = 1 \textwidth]{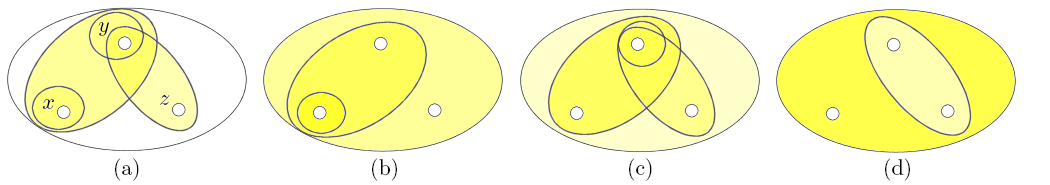}
\end{tabular}
\caption{\label{fig:example-l1-m} (a) The belief function (\ref{eq:example}); (b) its partial $L_1$ consistent approximation in $\mathcal{M}$ focussed on $\{x\}$; (c) partial approximation on $\{y\}$; (d) partial approximation on $\{z\}$.}
\end{figure}

Consider as an example the belief function $b$ with mass assignment
\begin{equation} \label{eq:example}
\begin{array}{c}
m_b(x)=0.2, \; m_b(y)=0.1, \; m_b(z)=0, \\ m_b(x,y)=0.4, \; m_b(x,z)=0, \; m_b(y,z)=0.3, \; m_b(\Theta)=0
\end{array}
\end{equation}
on the ternary frame $\Theta = \{x,y,z\}$.\\ Suppose that we seek the $L_1$ approximation of $b$ with core $\{x\}$. By Equation (\ref{eq:l1-m}) we get the following consonant b.f.:
\[
m'_b(x) = 0.2, \; m'_b(x,y)=0.4, \; m'_b(x,z)=0, \; m'_b(\Theta)=0.4.
\]
If, instead, we focus on $\{y\}$ or $\{z\}$ we get, respectively (see Figure \ref{fig:example-l1-m}):
\[
\begin{array}{cccc}
m'_b(y) = 0.1, & m'_b(x,y)=0.4, & m'_b(y,z)=0.3, & m'_b(\Theta)=0.2; \\ m'_b(z) = 0, & m'_b(x,z) = 0, & m'_b(y,z)=0.3, & m'_b(\Theta)=0.7.
\end{array}
\]
Note that the partial approximation on $\mathcal{CS}^z$ has $\{y,z\}$ as actual core, and that (b) and (d) are actually consonant (a special case of consistent b.f.), stressing the close relation between consonant and consistent approximation. Since $pl_b(x)=0.6$, $pl_b(y)=0.8$, $pl_b(z)=0.3$, by (\ref{eq:global-l1-m}) the global $L_1$ approximation of (\ref{eq:example}) is the partial approximation with core containing $\{y\}$. We can notice how the global approximation is visually much closer to the original b.f. than the partial ones, in terms of the structure of its focal elements.

\subsection{$L_\infty$ approximation} \label{sec:linf-m}

In the $L_\infty$ case $\displaystyle \| \vec{m}_b - \vec{m}_{cs} \|^{\mathcal{M}}_{L_\infty} = \max \Big \{ \max_{B \supseteq \{x\}, B \neq \Theta } |\beta(B)|, \max_{C \not \supset \{x\}} m_b(C) \Big \}$.\\
The $L_\infty$ norm of the difference vector is obviously minimized by $\{ \beta(B) \}$ such that: $|\beta(B)| \leq \max_{C \not \supset \{x\}} m_b(C)$ for all $B \supseteq \{x\}, B \neq \Theta$, i.e.,
\[
\displaystyle - \max_{C \not \supset \{x\}} m_b(C) \leq m_b(B) - m_{cs}(B) \leq \max_{C \not \supset \{x\}} m_b(C) \hspace{5mm} \forall \; B \supseteq \{x\}, B \neq \Theta.
\]
\begin{theorem} \label{the:linf-m}
Given an arbitrary belief function $b:2^\Theta \rightarrow [0,1]$ and an element $x \in \Theta$ of the frame, its partial $L_\infty$ consistent approximations $cs^x_{\mathcal{M},L_\infty}[m_b]$ with core containing $x$ in $\mathcal{M}$ are those whose mass values on all the subsets containing $x$ differ from the original ones by the maximum mass of the subsets not in the ultrafilter: for all $B \supset \{x\}$, $B \neq \Theta$
\begin{equation}\label{eq:linf-m}
m_b(B) - \max_{C \not \supset \{x\}} m_b(C) \leq m_{cs^x_{\mathcal{M},L_\infty}[m_b]}(B) \leq m_b(B) + \max_{C \not \supset \{x\}} m_b(C).
\end{equation}
\end{theorem}
Clearly this set of solutions can also include pseudo belief functions. Also, a comparison of Equations (\ref{eq:linf-m}) and (\ref{eq:l1-m}) shows that the barycenter of the partial $L_\infty$ approximations coincides with the partial $L_1$ approximation, reassigning all the mass outside the ultrafilter to the whole frame.

\subsubsection{Global approximation} Once again, the global $L_\infty$ consistent approximation in $\mathcal{M}$ coincides with the partial approximation (\ref{eq:linf-m}) at minimal distance from the original b.p.a. $\vec{m}_b$. The partial approximation focussed on $x$ has distance $\max_{C \not \supset \{x\}} m_b(C)$ from $\vec{m}_b$. The global $L_\infty$ approximation is therefore the (union of the) partial approximation(s) associated with the singleton(s) which minimize the maximal mass outside the ultrafilter:
\begin{equation} \label{eq:global-linf-m}
cs_{\mathcal{M},L_\infty}[m_b] = \bigcup_{\arg \min_{x} \max_{C \not \supset \{x\}} m_b(C)} cs^x_{\mathcal{M},L_\infty}[m_b].
\end{equation}
Global $L_\infty$ solutions are not totally unrelated to their $L_1$ peers, as maximizing the plausibility of the core as in (\ref{eq:global-l1-m}) involves minimizing the total mass outside the ultrafilter.

\subsubsection{Running example}

For the b.f. (\ref{eq:example}) of our running example, the maximal mass outside the ultrafilter in $x$ is
\[
\max_{C \not \supset x} m_b(C) = \max \big \{ m_b(y), m_b(z), m_b(y,z) \big \} = m_b(y,z) = 0.3,
\]
so that the set of $L_\infty$ consistent approximations with core containing $\{x\}$ is such that
\begin{equation} \label{eq:example-linf-m}
\left \{ \begin{array}{l} m_b(x) - m_b(y,z) \leq m'_b(x) \leq m_b(x) + m_b(y,z) \\ m_b(x,y) - m_b(y,z) \leq m'_b(x,y) \leq m_b(x,y) + m_b(y,z) \\ m_b(x,z) - m_b(y,z) \leq m'_b(x,z) \leq m_b(x,z) + m_b(y,z) \end{array} \right. \equiv \left \{ \begin{array}{l} -0.1 \leq m'_b(x) \leq 0.5 \\ 0.1 \leq m'_b(x,y) \leq 0.7 \\ -0.3 \leq m'_b(x,z) \leq 0.3. \end{array} \right.
\end{equation}
This set is clearly not entirely admissible. The same can be verified for partial approximations with cores containing $\{y\}$ and $\{z\}$, for which
\[
\max_{C \not \supset y} m_b(C) = m_b(x) = 0.2, \;\;\; \max_{C \not \supset z} m_b(C) = m_b(x,y) = 0.4.
\]
Therefore, the global $L_\infty$ approximations of (\ref{eq:example}) are the partial ones whose core contains $y$ (as it was the case for the global $L_1$ approximation).

\subsection{$L_2$ approximation} \label{sec:l2-m}

In order to find the $L_2$ consistent approximation(s) in $\mathcal{M}$ it is convenient to recall that the minimal $L_2$ distance between a point and a vector space is attained by the point of the vector space $V$ such that the difference vector is orthogonal to all the generators $\vec{g}_i$ of $V$:
\[
\arg \min_{\vec{q} \in V} \| \vec{p} - \vec{q} \|_2 = \hat{q} \in V : \langle \vec{p} -\hat{q}, \vec{g}_i \rangle = 0 \;\;\; \forall i
\]
whenever $\vec{p}\in\mathbb{R}^m$, $V = span(\vec{g}_i, i)$.

Instead of minimizing the $L_2$ norm of the difference vector $\| \vec{m}_b - \vec{m}_{cs} \|^{\mathcal{M}}_{L_2}$ we can just impose the orthogonality of the difference vector itself $\vec{m}_b - \vec{m}_{cs}$ and the subspace $\mathcal{CS}_\mathcal{M}^x$ associated with consistent mass functions focused on $\{x\}$. Clearly the generators of such linear space are the vectors in $\mathcal{M}$: $\vec{m}_B - \vec{m}_{\{x\}}$, for all $B \supsetneq \{x\}$.

\begin{theorem} \label{the:l2-m}
Consider an arbitrary belief function $b:2^\Theta \rightarrow [0,1]$ and an element $x \in \Theta$ of the frame. When using the ($N-2$)-dimensional representation of mass vectors (\ref{eq:dev-m}), its unique $L_2$ partial consistent approximation in $\mathcal{M}$ with core containing $x$ coincides with its partial $L_1$ approximation: $cs^x_{\mathcal{M},L_2}[m_b] = cs^x_{\mathcal{M},L_1}[m_b]$. When representing belief functions as mass
vectors $\vec{m}_b$ of $\mathbb{R}^{N-1}$ ($B=\Theta$ included)
\begin{equation} \label{eq:dev-m-N-1}
\vec{m}_b = \sum_{\emptyset \subsetneq B \subseteq \Theta} m_b(B) \vec{m}_B
\end{equation}
the partial $L_2$ approximation of $b$ is obtained by equally redistributing to each element of the ultrafilter $\{B \supseteq \{x\}\}$ an equal fraction of the mass of focal elements originally not in it:
\begin{equation}\label{eq:l2-m}
m_{cs^x_{\mathcal{M},L_2}[m_b]}(B) = m_b(B) + \frac{b(\{x\}^c)}{2^{|\Theta|-1}} \;\;\;\;\; \forall B \supseteq \{x\}.
\end{equation}
\end{theorem}
The partial $L_2$ approximation in $\mathbb{R}^{N-1}$ redistributes the mass equally to all the elements of the ultrafilter.

\subsubsection{Global approximation}

The global $L_2$ consistent approximation in $\mathcal{M}$ is, as usual, given by the partial approximation (\ref{eq:l2-m}) at minimal $L_2$ distance from $\vec{m}_b$. In the $N-2$ representation, by definition of $L_2$ norm in $\mathcal{M}$ (\ref{eq:norms-in-M}), the partial approximation focussed on $x$ has distance from $\vec{m}_b$
\[
(b(x^c))^2 + \sum_{B \not \supset \{x\}} ( m_b(B) )^2 = \Big ( \sum_{B \not \supset \{x\}} m_b(B) \Big)^2 + \sum_{B \not \supset \{x\}} ( m_b(B) )^2,
\]
which is minimized by the singleton(s) $\displaystyle \arg \min_{x} \sum_{B \not \supset \{x\}} (m_b(B))^2$. Therefore:
$\displaystyle
cs_{\mathcal{M},L_2}[m_b] = \bigcup_{\arg \min_{x} \sum_{B \not \supset \{x\}} (m_b(B))^2 } cs^x_{\mathcal{M},L_2}[m_b].
$

In the $N-1$-dimensional representation, instead,
\begin{equation}\label{eq:l2norm}
\begin{array}{l}
\displaystyle \sum_{B \supseteq \{x\}, B \neq \Theta} \Big [ m_b(B) - \Big ( m_b(B) + \frac{b(x^c)}{2^{|\Theta|-1}} \Big ) \Big ]^2 + \sum_{B \not \supset \{x\}}  (m_b(B))^2 = \\ \displaystyle \sum_{B \supseteq \{x\}, B \neq \Theta} \Big ( \frac{b(x^c)}{2^{|\Theta|-1}} \Big )^2 + \sum_{B \not \supset \{x\}}  (m_b(B))^2 = \frac{(\sum_{B \not \supset \{x\}} m_b(B))^2}{2^{|\Theta|-1}} + \sum_{B \not \supset \{x\}} (m_b(B))^2
\end{array}
\end{equation}
which is minimized by the same singleton(s). In any case, even though (in the $N-2$ representation) the partial $L_1$ and $L_2$ approximations coincide, the global approximations in general may fall on different components of the consonant complex.

\subsubsection{Running example}

\begin{figure}[ht!]
\centering
\begin{tabular}{c}
\includegraphics[width = 1 \textwidth]{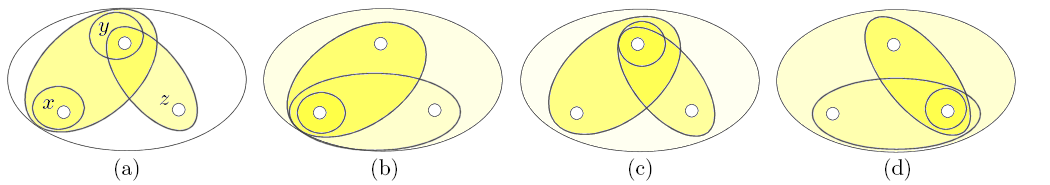}
\end{tabular}
\caption{\label{fig:example-l2-m} (a) The b.f. (\ref{eq:example}); (b) its partial $L_2$ consistent approximation in $\mathcal{M}$, focussed on $\{x\}$; (c) $L_2$ partial approximation on $\{y\}$; (d) $L_2$ partial approximation on $\{z\}$.}
\end{figure}

For the usual example b.f. (\ref{eq:example}), by Equation (\ref{eq:l2-m}) the $L_2$ partial approximations with cores containing $x$, $y$ and $x$ respectively have mass assignments:
\[
\begin{array}{llll}
m'_b(x) = 0.3, & m'_b(x,y) = 0.5, & m'_b(x,z) = 0.1, & m'_b(\Theta) = 0.1 \\ m'_b(y) = 0.15, & m'_b(x,y) = 0.45, & m'_b(y,z) = 0.35, & m'_b(\Theta) = 0.05; \\ m'_b(z) = 0.175, & m'_b(x,z) = 0.175, & m'_b(y,z) = 0.475, & m'_b(\Theta) = 0.175,
\end{array}
\]
since $b(x^c) = 1 - pl_b(x) = 0.4$, $b(x^c) = 1 - pl_b(x) = 0.4$, $b(x^c) = 1 - pl_b(x) = 0.4$ and $2^{|\Theta|-1} = 4$ (see Figure \ref{fig:example-l2-m}).

Note that, unlike the corresponding $L_1$ approximation, the partial $L_2$ approximation on $\{z\}$ has indeed the singleton as core. Also, as $L_2$ approximation redistributes mass evenly to all the focal elements in the desired ultrafilter, all partial approximations contain all possible f.e.s. As for the global solution
\[
\begin{array}{ll}
(b(x^c))^2 = 0.16, & \displaystyle \sum_{B \not \supset x} (m_b(B))^2 = 0.1^2 + 0.3^2 = 0.01 + 0.09 = 0.1, \\ (b(y^c))^2 = 0.04, & \displaystyle \sum_{B \not \supset y} (m_b(B))^2 = 0.2^2 = 0.04, \\ (b(z^c))^2 = 0.49, & \displaystyle \sum_{B \not \supset z} (m_b(B))^2 = 0.2^2 + 0.1^2 + 0.4^2 = 0.04 + 0.01 + 0.16 = 0.21, \\
\end{array}
\]
and the $L_2$ norm (\ref{eq:l2norm}) is minimized, once again, by the partial solution on $\{y\}$ (as visually confirmed by Figure \ref{fig:example-l2-m}).

\section{Calculation: approximation in the belief space} \label{sec:approximation-belief}

\subsection{$L_1$/$L_2$ approximations} \label{sec:l1l2}

We have seen that in the mass space (at least in its $N-2$ representation, Theorem \ref{the:l2-m}) the $L_1$ and $L_2$ approximations coincide. This is true in the belief space in the general case as well. We will gather some intuition on the general solution by considering first the slightly more complex case of a ternary frame: $\Theta = \{x,y,z\}$. In this Section we will use the notation:
\[
\vec{cs} = \sum_{B \supseteq \{x\}} m_{cs}(B) \vec{b}_B, \;\;\; \vec{b} = \sum_{B \subsetneq \Theta} m_b(B) \vec{b}_B.
\]

In the case of an arbitrary frame a cs.b.f. $cs \in \mathcal{CS}_\mathcal{B}^x$ is a solution of the $L_2$ approximation problem if, again, $\vec{b}-\vec{cs}$ is orthogonal to all generators $\{ \vec{b}_B - \vec{b}_\Theta = \vec{b}_B, \{ x \} \subseteq B \subsetneq \Theta \}$ of $\mathcal{CS}_\mathcal{B}^x$: $\langle \vec{b}-\vec{cs}, \vec{b}_B \rangle = 0$ $\forall B : \{ x \} \subseteq B \subsetneq \Theta$. As $\vec{b}-\vec{cs} = \sum_{A \subsetneq \Theta} (m_b(A) - m_{cs}(A)) \vec{b}_A$ we get
\begin{equation} \label{eq:system-l2-general}
\left \{ \sum_{A \supseteq \{x\}} \beta(A) \langle \vec{b}_A,\vec{b}_B \rangle + \sum_{A \not \supset \{x\}} m_b(A) \langle \vec{b}_A, \vec{b}_B \rangle = 0 \right . \hspace{5mm} \forall B : \{ x \} \subseteq B \subsetneq \Theta,
\end{equation}
where once again $\beta(A)=m_b(A) - m_{cs}(A)$. In the $L_1$ case, the minimization problem is (using again the notation $\vec{cs} = \sum_{B \supseteq \{x\}} m_{cs}(B) \vec{b}_B$)
\[
\begin{array}{lll}
& & \displaystyle \arg\min_{m_{cs}(.)} \bigg \{ \sum_{B \supseteq \{x\}} \bigg | \sum_{A \subseteq B} m_b(A) - \sum_{A \subseteq B, A \supseteq \{x\}} m_{cs}(A) \bigg | \bigg \} \\ & = & \displaystyle \arg\min_{\beta(.)} \bigg \{ \sum_{B \supseteq \{x\}} \bigg | \sum_{A \subseteq B, A \supseteq \{x\}} \beta(A) + \sum_{A \subseteq B, A \not\supset \{x\}} m_b(A) \bigg | \bigg \}
\end{array}
\]
which is clearly solved by setting all addenda to zero, obtaining
\begin{equation} \label{eq:system-l1-general}
\left \{ \sum_{A \subseteq B, A \supseteq \{x\}} \beta(A) + \sum_{A \subseteq B, A \not\supset \{x\}} m_b(A) = 0 \hspace{5mm} \forall B : \{ x \} \subseteq B \subsetneq \Theta. \right .
\end{equation}

An interesting fact emerges when comparing the linear systems for $L_1$ and $L_2$ in the ternary case $\Theta = \{x,y,x\}$:
\begin{equation} \label{eq:system-comparison}
\begin{array}{l}
\left\{ \begin{array}{l}
\begin{array}{l} 3 \beta(x) + \beta(x,y) + \beta(x,z) + m_b(y) + m_b(z) = 0 \end{array} \\ \beta(x) + \beta(x,y) + m_b(y) = 0 \\ \beta(x) + \beta(x,z) + m_b(z) = 0
\end{array} \right.
\\
\left\{ \begin{array}{l}
\beta(x) = 0 \\ \beta(x) + \beta(x,y) + m_b(y) = 0 \\ \beta(x) + \beta(x,z) + m_b(z) = 0.
\end{array} \right.
\end{array}
\end{equation}
The solution is the same for both, due to the fact that the second linear system is obtained from the first one by a linear transformation of rows. We just need to replace the first equation $e_1$ in the first system with the difference: $e_1 \mapsto e_1 - e_2 - e_3$. This holds in the general case, too.
\begin{lemma} \label{lem:coeff}
$\displaystyle \sum_{C \supseteq B} \langle \vec{b}_C,\vec{b}_A \rangle (-1)^{|C \setminus B|} = 1$ whenever $A \subseteq B$, 0 otherwise.
\end{lemma}
\begin{corollary}
The linear system (\ref{eq:system-l2-general}) can be reduced to the system (\ref{eq:system-l1-general}) through the following linear transformation of rows:
\begin{equation} \label{eq:trans}
row_B \mapsto \sum_{C \supseteq B} row_C (-1)^{|C \setminus B|}.
\end{equation}
\end{corollary}
\emph{Proof}.
If we apply the linear transformation (\ref{eq:trans}) to the system (\ref{eq:system-l2-general}) we get
\[
\begin{array}{lll}
& \displaystyle \sum_{C \supseteq B} \bigg [ \sum_{A \supseteq \{x\}} \beta(A) \langle \vec{b}_A, \vec{b}_C \rangle + \sum_{A \not \supset \{x\}} m_b(A) \langle \vec{b}_A, \vec{b}_C \rangle \bigg ] (-1)^{|C \setminus B|} \\ = & \displaystyle \sum_{A \supseteq \{x\}} \beta(A) \sum_{C \supseteq B} \langle \vec{b}_A, \vec{b}_C \rangle (-1)^{|C \setminus B|} + \sum_{A \not \supset \{x\}} m_b(A) \sum_{C \supseteq B} \langle \vec{b}_A, \vec{b}_C \rangle (-1)^{|C \setminus B|}
\end{array}
\]
$\forall B : \{ x \} \subseteq B \subsetneq \Theta$. Therefore by Lemma \ref{lem:coeff} we get
\[
\bigg \{\sum_{A \supseteq \{x\}, A \subseteq B} \beta(A) + \sum_{A \not\supset \{x\}, A \subseteq B} m_b(A) = 0 \hspace{5mm} \forall B : \{ x \} \subseteq B \subsetneq \Theta,
\]
i.e., the system of equations (\ref{eq:system-l1-general}). $\Box$

\subsubsection{Form of the solution}

To obtain both the $L_2$ and the $L_1$ consistent approximations of $b$ it then suffices to solve the system (\ref{eq:system-l1-general}) associated with the $L_1$ norm.
\begin{theorem}
The unique solution of the linear system (\ref{eq:system-l1-general}) is: $\beta(A) = - m_b(A \setminus \{x\})$ for all $A : \{ x \} \subseteq A \subsetneq \Theta$.
\end{theorem}
We can prove it by simple substitution, as system (\ref{eq:system-l1-general}) becomes:
\[
\begin{array}{l}
\displaystyle - \sum_{B \subseteq A, B \supseteq \{x\}} m_b(B \setminus \{x\}) + \sum_{B \subseteq A, B \not \supset \{x\}} m_b(B) \\ = \displaystyle - \sum_{C \subseteq A \setminus \{x\}} m_b(C) + \sum_{B \subseteq A, B \not \supset \{x\}} m_b(B) = 0.
\end{array}
\]
Therefore, according to what discussed in Section \ref{sec:approximation-complex}, the partial consistent approximations of $b$ on the maximal component $\mathcal{CS}_\mathcal{B}^x$ of the consistent complex have b.p.a.
\[
m_{cs^x_{\mathcal{B},L_1}[b]}(A) = m_{cs^x_{\mathcal{B},L_2}[b]}(A) = m_b(A) - \beta(A) = m_b(A) + m_b(A \setminus \{x\})
\]
for all events $A$ such that $\{x\} \subseteq A \subsetneq \Theta$. As for the mass of $\Theta$, by normalization: $m_{cs^x_{\mathcal{B},L_1/L_2}[b]}(\Theta) =$
\[
\begin{array}{lll}
& = & \displaystyle 1 - \sum_{\{x\} \subseteq A \subsetneq \Theta} m_{cs^x_{\mathcal{B},L_1/L_2}[b]}(A) = 1 - \sum_{\{x\} \subseteq A \subsetneq \Theta} \Big ( m_b(A) + m_b(A \setminus \{x\}) \Big ) \\ & = & \displaystyle 1 - \sum_{\{x\} \subseteq A \subsetneq \Theta} m_b(A) - \sum_{\{x\} \subseteq A \subsetneq \Theta} m_b(A \setminus \{x\}) \\ & = & \displaystyle 1 - \sum_{A \neq \Theta, \{x\}^c} m_b(A) = m_b(\{x\}^c) + m_b(\Theta)
\end{array}
\]
as all events $B\not\supset \{x\}$ can be written as $B = A \setminus \{x\}$ for $A = B \cup \{x\}$.
\begin{corollary} \label{cor:l1l2-b}
\[
m_{cs^x_{\mathcal{B},L_1}[b]}(A) = m_{cs^x_{\mathcal{B},L_2}[b]}(A) = m_b(A) + m_b(A \setminus \{x\})
\]
$\forall x \in \Theta$, and for all $A$ s.t. $\{x\} \subseteq A \subseteq \Theta$.
\end{corollary}

\subsubsection{Global $L_1$ and $L_2$ approximations} \label{sec:global-l1-l2}

To find the consistent approximation of $b$ we need to work out which of the partial approximations $cs^x_{L_{1/2}}[b]$ has minimal distance from $b$.
\begin{theorem} \label{the:global-l1}
Given an arbitrary belief function $b:2^\Theta \rightarrow [0,1]$, its global $L_1$ consistent approximation in the belief space is its partial approximation associated with the singleton:
\begin{equation} \label{eq:xhat-l1}
\arg\min_x \Big \{ \sum_{A \subseteq \{x\}^c} b(A) , x \in\Theta \Big \}.
\end{equation}
\end{theorem}
In the binary case ($\Theta = \{x,y\}$) the optimal singleton cores (\ref{eq:xhat-l1}) simplify as:
\[
\arg\min_x \sum_{A \subseteq \{x\}^c} b(A) = \arg\min_x m_b(\{x\}^c) = \arg\max_x pl_b(x),
\]
and the global approximation falls on the component of the consistent complex associated with the element of \emph{maximal plausibility}. Unfortunately, in the case of an arbitrary frame $\Theta$ the element (\ref{eq:xhat-l1}) is not necessarily the maximal plausibility element:
\[
\arg\min \Big \{ \sum_{A \subseteq \{x\}^c} b(A) , x \in\Theta \Big \} \neq \arg\max \{ pl_b(x) , x \in\Theta \big \},
\]
as a simple counterexample can prove. As for the $L_2$ case:
\begin{theorem} \label{the:global-l2}
Given an arbitrary belief function $b:2^\Theta \rightarrow [0,1]$, its global $L_2$ consistent approximation in the belief space is its partial approximation associated with the singleton:
\begin{equation} \label{eq:xhat-l2}
\displaystyle \arg\min_x \Big \{ \sum_{A \subseteq \{x\}^c} \big ( b(A) \big )^2 , x \in\Theta \Big \}.
\end{equation}
\end{theorem}
Once again, in the binary case the optimal singleton cores (\ref{eq:xhat-l2}) specialize as:
\[
\arg\min_x \sum_{A \subseteq \{x\}^c} (b(A))^2 = \arg\min_x (m_b(\{x\}^c))^2 = \arg\max_x pl_b(x)
\]
and the global approximation for $L_2$ also falls on the component of the consistent complex associated with the element of maximal plausibility, while this is not generally true for an arbitrary frame.

\subsubsection{Running example} \label{sec:example-belief-l1-l2}

\begin{figure}[ht!]
\centering
\begin{tabular}{c}
\includegraphics[width = 1 \textwidth]{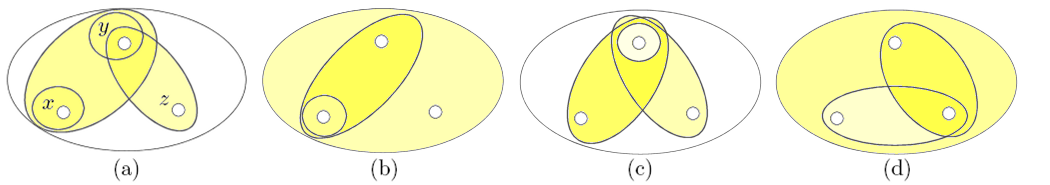}
\end{tabular}
\caption{\label{fig:example-l1-l2-b} (a) The belief function (\ref{eq:example}); (b) its partial $L_1/L_2$ consistent approximation in $\mathcal{B}$, focussed on $\{x\}$; (c) partial $L_1/L_2$ partial approximation on $\{y\}$; (d) partial $L_1/L_2$ partial approximation on $\{z\}$.}
\end{figure}

According to Equation (\ref{eq:l1l2b}), for the belief function (\ref{eq:example}) of our running example the partial $L_1$/$L_2$ consistent approximations in $\mathcal{B}$ with core containing $\{x\}$, $\{y\}$ and $\{z\}$, respectively, have mass assignments (see Figure \ref{fig:example-l1-l2-b}):
\begin{equation} \label{eq:example-l2-b}
\begin{array}{llll}
m'_b(x) = 0.2, & m'_b(x,y) = 0.5, & m'_b(x,z) = 0, & m'_b(\Theta) = 0.3; \\ m'_b(y) = 0.1, & m'_b(x,y) = 0.6, & m'_b(y,z) = 0.3, & m'_b(\Theta) = 0; \\ m'_b(z) = 0, & m'_b(x,z) = 0.2, & m'_b(y,z) = 0.4, & m'_b(\Theta) = 0.4,
\end{array}
\end{equation}
We can note that kind of approximation tends to penalize the smallest, singleton element of the ultrafilter $\{B \supseteq x\}$: this is due to the fact that $m'_b(x) = m_b(x)$, while the masses of all the other elements of the filter are increased. As for global approximations we have, for $L_1$:
\[
\begin{array}{l}
\sum_{A \subseteq \{x\}^c} b(A) = b(y) + b(z) + b(y,z) = 0.1 + 0 + 0.4 = 0.5, \\ \sum_{A \subseteq \{y\}^c} b(A) = b(x) + b(z) + b(x,z) = 0.2 + 0 + 0.2 = 0.4, \\ \sum_{A \subseteq \{z\}^c} b(A) = b(x) + b(y) + b(x,y) = 0.2 + 0.1 + 0.7 = 1,
\end{array}
\]
which is minimized by the partial solution in $y$. For $L_2$:
\[
\begin{array}{l}
\sum_{A \subseteq \{x\}^c} \big ( b(A) \big )^2 = 0.01 + 0.16 = 0.17, \\ \sum_{A \subseteq \{y\}^c} \big ( b(A) \big )^2 = 0.04 + 0.04 = 0.08, \\ \sum_{A \subseteq \{ z \}^c} \big ( b(A) \big )^2 = 0.04 + 0.01 + 0.49 = 0.54,
\end{array}
\]
which is also minimized by the same partial solution. This is not necessarily the case, in general.

\subsection{$L_\infty$ consistent approximation} \label{sec:linfty}

While the partial $L_1$- and $L_2$-consistent approximations in the belief space are pointwise and coincide, the set of partial $L_\infty$-approximations for each component $\mathcal{CS}_\mathcal{B}^x$ of the consistent complex form a polytope whose center of mass is exactly equal to $cs_{\mathcal{B},L_1}^x[b] = cs_{\mathcal{B},L_2}^x[b]$.

\subsubsection{General solution}

By applying the expression (\ref{eq:lp-b}) of the $L_\infty$ norm to the difference vector $\vec{b} - \vec{cs}$ in the belief space, where $cs$ is a consistent b.f. with core containing $x$, we obtain $\max_{A \subsetneq \Theta} \big \{ \big | \sum_{B \subseteq A} m_b(B) - \sum_{B \subseteq A, B \supseteq \{x\}} m_{cs}(B) \big | \big \}$. Therefore, the partial $L_\infty$ consistent approximation with core containing $x$ is:
\[
cs^x_{\mathcal{B},L_\infty} [b] = \arg\min_{m_{cs}(.)} \max_{A \subsetneq \Theta} \bigg \{ \bigg | \sum_{B \subseteq A} m_b(B) - \sum_{B \subseteq A, B \supseteq \{x\}} m_{cs}(B) \bigg | \bigg \}.
\]
Again, $\max_{A \subsetneq \Theta}$ has as lower limit the value associated with the largest norm which does not depend on $m_{cs}(.)$, i.e.,
\[
\max_{A \subsetneq \Theta} \bigg \{ \bigg | \sum_{B \subseteq A} m_b(B) - \sum_{B \subseteq A, B \supseteq \{x\}} m_{cs}(B) \bigg | \bigg \} \geq b(\{x\}^c)
\]
or equivalently $\displaystyle \max_{A \subsetneq \Theta} \bigg \{ \bigg | \sum_{B \subseteq A, B \supseteq \{x\}} \beta(B) + \sum_{B \subseteq A, B \not \supset \{x\}} m_b(B) \bigg | \bigg \} \geq b(\{x\}^c).$\\
In the above constraint only the expressions associated with $B \supseteq \{x\}$ contain variable terms. Therefore the desired optimal values of the variables $\{\beta(B)\}$ are such that:
\begin{equation} \label{eq:system-linf-general}
\left \{ \bigg | \sum_{B \subseteq A, B \supseteq \{x\}} \beta(B) + \sum_{B \subseteq A, B \not\supset \{x\}} m_b(B) \bigg | \leq b(\{x\}^c) \hspace{5mm} \forall A: \{x\} \subseteq A \subsetneq \Theta. \right .
\end{equation}
After introducing the change of variables
\begin{equation} \label{eq:change}
\gamma(A) \doteq \sum_{B \subseteq A, B \supseteq \{x\}} \beta(B)
\end{equation}
system (\ref{eq:system-linf-general}) trivially reduces to
$\displaystyle \Big \{ \Big | \gamma(A) + \sum_{B \subseteq A, B \not\supset \{x\}} m_b(B) \Big | \leq b(\{x\}^c)$ for all $A$ such that $\{x\} \subseteq A \subsetneq \Theta$, whose solution
\begin{equation} \label{eq:gamma-solution}
-b(x^c) - \sum_{B \subseteq A, B \not\supset \{x\}} m_b(B) \leq \gamma(A) \leq b(x^c) - \sum_{B \subseteq A, B \not\supset \{x\}} m_b(B)
\end{equation}
defines a high-dimensional ``rectangle" in the space of the solutions $\{\gamma(A),\{x\} \subseteq A \subsetneq \Theta\}$.\\ In the mass assignment $m_{cs}(.)$ of the desired approximations, as we can clearly see in the running ternary example, the solution set is polytope whose vertices do not appear to have straightforward interpretations. On the other hand, the barycenter of this polytope is easy to compute and interpret.

\subsubsection{Barycenter of the $L_\infty$ solution and global approximation}

The center of mass of the set of solutions (\ref{eq:gamma-solution}) to the $L_\infty$ consistent approximation problem is clearly $\displaystyle \gamma(A) = - \sum_{B \subseteq A, B \not\supset \{x\}} m_b(B)$, $\{x\} \subseteq A \subsetneq \Theta$, which reads in the space of the variables $\{\beta(A),\{x\} \subseteq A \subsetneq \Theta\}$ as
\[
\bigg \{ \sum_{B \subseteq A, B \supseteq \{x\}} \beta(B) = - \sum_{B \subseteq A, B \not\supset \{x\}} m_b(B), \hspace{10mm} \{x\} \subseteq A \subsetneq \Theta.
\]
But this is exactly the linear system (\ref{eq:system-l1-general}) which determines the $L_1/L_2$ consistent approximation $cs_{\mathcal{B},L_{1/2}}^x[b]$ of $b$ onto $\mathcal{CS}_\mathcal{B}^x$.\\ Besides, the $L_\infty$ distance between $b$ and $\mathcal{CS}_\mathcal{B}^x$ is minimal for the element $x$ which minimizes $\| \vec{b}-\vec{cs}^x_{\mathcal{B},L_\infty} \|_\infty = b(\{x\}^c)$. In conclusion,
\begin{theorem}
Given a belief function $b:2^\Theta \rightarrow [0,1]$, and an element of its frame $x \in \Theta$, its partial $L_1/L_2$ approximation onto any given component $\mathcal{CS}_\mathcal{B}^x$ of the consistent complex $\mathcal{CS}_\mathcal{B}$ in the belief space is also the geometric barycenter of the set its $L_\infty$ consistent approximations on the same component. Its global $L_\infty$ consistent approximations in $\mathcal{B}$ form the union of the partial $L_\infty$ approximations associated with the maximal plausibility element(s) $x \in \Theta$.
\end{theorem}

\subsection{Running example}

For the usual belief function (\ref{eq:example}) on the ternary frame, the set of partial $L_\infty$ solutions (\ref{eq:gamma-solution}) focussed on $x$ becomes:
\[
\begin{array}{ccc}
b(x) - b(x^c) \leq & m_{cs}(x) & \leq b(x) + b(x^c), \\ b(x,y) - b(x^c) \leq & m_{cs}(x) + m_{cs}(x,y) & \leq b(x,y) + b(x^c), \\ b(x,z) - b(x^c) \leq & m_{cs}(x) + m_{cs}(x,z) & \leq b(x,z) + b(x^c),
\end{array}
\]
whose $2^3=8$ vertices (represented as vectors $[m_{cs}(x),m_{cs}(x,y),m_{cs}(x,z),m_{cs}(\Theta)]'$)
\[
\begin{array}{lcl}
\scriptsize \big [ b(x) - b(x^c), & b(x,y) - b(x), & b(x,z) - b(x), \\ & & 1 + b(x) + b(x^c) - b(x,y) - b(x,z) \big ], \\
\big [ b(x) - b(x^c), & b(x,y) - b(x), & b(x,z) - b(x) + 2 b(x^c), \\ & & 1 + b(x) - b(x,y) - b(x,z) - b(x^c) \big ],
\end{array}
\]
\[
\begin{array}{lcl}
\big [ b(x)-b(x^c), & b(x,y) - b(x) + 2b(x^c), & b(x,z) - b(x), \\ & & 1 + b(x) - b(x,y) - b(x,z) - b(x^c) \big ], \\ \big [ b(x)-b(x^c), & b(x,y) - b(x) + 2b(x^c), & b(x,z) - b(x) + 2b(x^c), \\ & & 1 + b(x) - b(x,y) - b(x,z) - 3 b(x^c) \big ], \\ \big [ b(x)+ b(x^c), & b(x,y) - b(x) - 2b(x^c), & b(x,z) - b(x) - 2b(x^c), \\ & & 1 + b(x) - b(x,y) - b(x,z) + 3 b(x^c) \big ], \\ \big [ b(x) + b(x^c), & b(x,y) - b(x) - 2b(x^c), & b(x,z) - b(x), \\ & & 1 + b(x) - b(x,y) - b(x,z) + b(x^c) \big ], \\ \big [ b(x) + b(x^c), & b(x,y) - b(x), & b(x,z) - b(x) - 2 b(x^c), \\ & & 1 + b(x) - b(x,y) - b(x,z) + b(x^c) \big ], \\ \big [ b(x) + b(x^c), & b(x,y) - b(x), & b(x,z) - b(x), \\ & & 1 + b(x) - b(x,y) - b(x,z) - b(x^c) \big ],
\end{array}
\]
do not appear to be particularly meaningful. Their barycenter, instead,
\begin{equation} \label{eq:example-barycenter-linf-b}
\begin{array}{ll}
& \big [ b(x), b(x,y) - b(x), b(x,z) - b(x), 1 + b(x) - b(x,y) - b(x,z)  \big ] \\ = & \big [ m_b(x), m_b(x,y) + m_b(y), m_b(x,z) + m_b(z), m_b(y,z) + m_b(\Theta) \big ]
\end{array}
\end{equation}
clearly coincides with the focussed consistent approximation on $x$. The same holds for the barycenters of the $L_\infty$ approximations focussed on $y$ and $z$ (see Equation (\ref{eq:example-l2-b})).

\section{Discussion} \label{sec:versus}

Let us conclude by discussing the interpretation of the consistent approximations derives in the mass and belief space, and by comparing their features and geometry, with the help of the usual ternary example.

\subsection{Approximation in the mass space and general imaging in belief revision}

As it is the case for geometric conditioning in the mass space \cite{cuzzolin10brest}, consistent approximations in the mass space can be interpreted as a generalization of Lewis' \emph{imaging} approach to belief revision, originally formulated in the context of probabilities \cite{lewis76}. The idea behind imaging is that, upon observing that some state $x\in\Theta$ is impossible, you transfer the probability initially assigned to $x$ completely towards the remaining state you deem the most similar to $x$ \cite{perea09amodel}. Peter G\"{a}rdenfors \cite{Gardenfors} extended Lewis' idea by allowing a fraction $\lambda_i$ of the probability of such state $x$ to be re-distributed to all remaining states $x_i$ ($\sum_i \lambda_i = 1$).\\ In the case of partial consistent approximation of belief functions, the mass $m_b(A)$ of each focal element not in the desired ultrafilter $\{A \supseteq \{x\}\}$ should be re-assigned to the ``closest" focal element in the latter. The partial $L_1$ consistent approximation in $\mathcal{M}$ amounts to admitting ignorance about the closeness of focal elements, and re-assigning all the mass outside the filter to the whole frame, therefore increasing the uncertainty of the state. If such ignorance is expressed by assigning instead equal weight $\lambda(A)$ to each $A \in \mathcal{C}$, the resulting partial consistent approximation is the unique partial $L_2$ approximation, the barycenter of the polytope of $L_1$ partial approximations.

\subsection{Partial L1/L2 approximations as focused consistent transformations}

\begin{figure}[ht!]
\centering
\includegraphics[width = 0.5 \textwidth]{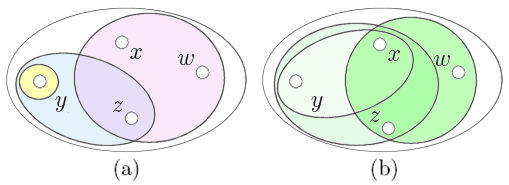}
\caption{\label{fig:example-partial} A belief function (a) and its $L_{1/2}$ consistent approximation in $\mathcal{B}$ with core $\{x\}$ (b).}
\end{figure}

The expression of the the basic probability assignment of the $L_1$/$L_2$ consistent approximations of $b$ (Corollary \ref{cor:l1l2-b}) is simple and elegant. It also has a straightforward interpretation: to get a consistent belief function focused on a singleton $x$, the mass contribution of all events $B$ such that $B \cup \{x\} = A$ coincide is assigned indeed to $A$. But there are just two such events: $A$ itself, and $A \setminus \{x\}$.\\ The partial consistent approximation of a belief function on a frame $\Theta = \{x,y,z,w\}$ with core $\{x\}$ is illustrated in Figure \ref{fig:example-partial}.
The b.f. with focal elements $\{y\}$, $\{y,z\}$, and $\{x,z,w\}$ is transformed by the map
\[
\begin{array}{lll}
\{y\} & \mapsto & \{x\} \cup \{y\} = \{x,y\}, \\ \{y,z\} & \mapsto & \{x\} \cup \{y,z\} = \{x,y,z\}, \\ \{x,z,w\} & \mapsto & \{x\} \cup \{x,z,w\} = \{x,z,w\}
\end{array}
\]
into the consistent b.f. with focal elements $\{x,y\}$, $\{x,y,z\}$, and $\{x,z,w\}$ and the same b.p.a. The partial solutions to the $L_1/L_2$ consistent approximation problem turn out to be related to the classical \emph{inner consonant approximations} of a b.f. $b$, i.e., the set of consonant belief functions $co$ such that
\[
co (A)\geq b(A) \hspace{5mm} \forall A \subseteq \Theta
\]
(or equivalently $pl_{co}(A)\leq pl_b(A)$ $\forall A$). Dubois and Prade \cite{dubois90} proved indeed that such an approximation exists iff $b$ is consistent. However, when $b$ is \emph{not} consistent a ``focused consistent transformation" can be applied to get a new belief function $b'$ such that
\begin{equation} \label{eq:l1l2b}
\begin{array}{ccc}
m_{b'}(A\cup x_i) = m_b(A) & & \forall A\subseteq\Theta
\end{array}
\end{equation}
and $x_i$ is the element of $\Theta$ with highest plausibility.\\ Clearly then, the results of Theorem and Corollary say that the $L_1$/$L_2$ consistent approximation onto each component $\mathcal{CS}_\mathcal{B}^x$ of $\mathcal{CS}_\mathcal{B}$ generates the consistent transformation focused on $x$.

\subsection{Improper solutions and unnormalized belief functions}

In some cases, \emph{improper} partial solutions (in the sense that they potentially include negative mass assignments) can be generated by the $L_p$ minimization process. This situation is not entirely new. For instance, outer consonant approximations \cite{dubois90} also include infinitely many improper solutions: nevertheless, only the subset of acceptable solutions is retained. As in the consonant case \cite{cuzzolin11isipta-consonant}, the set of all (admissible and not) solutions is typically much simpler to describe geometrically, in terms of simplices or polytopes. Computing the set of \emph{proper} approximations in all cases requires significant further effort, which for reasons of clarity and length we reserve for the near future.\\
Additionally, in this work only ``normalized" belief functions, i.e., b.f.s whose mass of the empty set is nil, are considered. Unnormalized b.f.s, however, play an important role in the TBM \cite{ubf} as the mass of the empty set is an indicator of conflicting evidence. The analysis of the unnormalized case is also left to future work for lack of sufficient space here.

\subsection{Mass- versus belief- consistent approximations}

Summarizing, in the mass space:
\begin{itemize}
\item
the partial $L_1$ consistent approximation focussed on a certain element $x$ of the frame is obtained by reassigning all the mass $b(x^c)$ outside the filter to $\Theta$;
\item
the global approximation is associated, as expected, with cores containing the maximal plausibility element(s) of $\Theta$;
\item
the $L_\infty$ approximation generates a ``rectangle" of partial approximations, with barycenter in the $L_1$ partial approximation;
\item
the corresponding global approximations spans the component(s) focussed on the element(s) $x$ such that $\max_{B \not \supset x} m_b(B)$ is minimal;
\item
the $L_2$ partial approximation coincides with the $L_1$ one in the $N-2$ representation;
\item
in the $N-1$ representation the $L_2$ partial approximation reassigns the mass outside the desired filter ($b(x^c)$) to each element of the filter focussed on $x$ on equal basis;
\item
global approximations in the $L_2$ case are of more difficult interpretation.
\end{itemize}
In the belief space:
\begin{itemize}
\item
partial $L_1$/$L_2$ approximations coincide on each component of the consistent complex;
\item
such partial $L_1$/$L_2$ approximation turns out to be the consistent transformation \cite{dubois90} focused on the considered element of the frame: for all events $A$ such that $\{x\} \subseteq A \subseteq \Theta$
\[
m_{cs^x_{\mathcal{B},L_1}[b]}(A) = m_{cs^x_{\mathcal{B},L_2}[b]}(A) = m_b(A) + m_b(A \setminus \{x\});
\]
\item
the corresponding global solutions have not in general as core the maximal plausibility element, and may lie in general on different components of $\mathcal{CS}$; the $L_1$ global consistent approximation is associated with the singleton(s) $x \in \Theta$ such that: $\hat{x} = \arg \min_{x} \sum_{A \subseteq \{x\}^c} b(A)$, while the $L_2$ global approximation is associated with $\hat{x} = \arg \min_{x} \sum_{A \subseteq \{x\}^c} (b(A))^2$ which do not appear to have simple epistemic interpretations;
\item
the set of partial $L_\infty$ solutions form a polytope on each component of the consistent complex, whose center of mass lies on the partial $L_1$/$L_2$ approximation;
\item
the global $L_\infty$ solutions fall on the component(s) associated with the maximal plausibility element(s), and their center of mass, when such element is unique, is the consistent transformation focused on the maximal plausibility singleton \cite{dubois90}.
\end{itemize}

By comparing the behavior of the two classes of approximations we can notice a general pattern. Approximations in both mass and belief space reassign the mass outside the filter focussed on $x$, in different ways. However, mass consistent approximations reassign this mass $b(x^c)$ either to no focal element in the filter (i.e., to $\Theta$) or to all on an equal basis. They do not distinguish between focal elements w.r.t. their set-theoretic relationships with subsets $B \not \supset x$ outside the filter. In opposition, approximations in the belief space do distinguish them according to the focussed consistent transformation principle.

\subsection{Comparison on a ternary example}

\begin{figure}[ht!]
\vspace{-3mm}
\begin{center}
\includegraphics[width=0.45\textwidth]{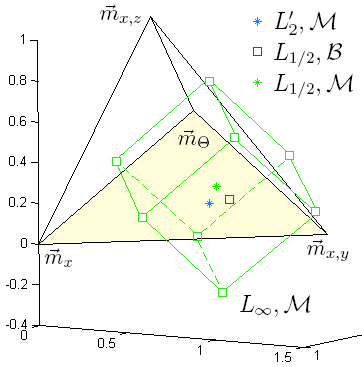}
\end{center}
\vspace{-5mm}
\caption{The simplex (solid black tetrahedron) $Cl(\vec{m}_x,\vec{m}_{x,y},\vec{m}_{x,z},\vec{m}_\Theta)$ of consistent belief functions focussed on $x$, and the related $L_p$ partial consistent approximations of a generic b.f. on $\Theta = \{ x,y,z \}$.}
\label{fig:example} \vspace{-3mm}
\end{figure}

To conclude, let us graphically illustrate the different approximations for the belief function of the running ternary example. Figure \ref{fig:example} illustrates the different partial consistent approximations in the simplex $Cl(\vec{m}_x,\vec{m}_{x,y},\vec{m}_{x,z},\vec{m}_\Theta)$ of consistent belief functions focussed on $x$ in a ternary frame, for the belief function (\ref{eq:example}). This is the solid black tetrahedron of Figure \ref{fig:example}.

The set of partial $L_\infty$ approximations in $\mathcal{M}$ (\ref{eq:example-linf-m}) is represented by the green cube in the Figure. As expected, it does not entirely fall inside the tetrahedron of admissible consistent belief functions. Its barycenter (the green star) coincides with the $L_1$ partial consistent approximation in $\mathcal{M}$. The $L_2$ approximation in $\mathcal{M}$ does also coincide, as expected, with the $L_1$ approximation. There seems to be a strong case for the latter approximation, whose natural interpretation in terms of mass assignment is the following: all the mass outside the desired ultrafilter $\{B \supseteq x\}$ is reassigned to $\Theta$, increasing the overall uncertainty of the belief state.\\ The $L_2,\mathcal{M}$ partial approximation in the $N-1$ representation is distinct from the previous ones, but still falls inside the polytope of $L_\infty$ partial approximations (green cube) and is admissible, as it falls in the interior of the simplicial component $Cl(\vec{m}_x,\vec{m}_{x,y},\vec{m}_{x,z},\vec{m}_\Theta)$. It itself possesses a strong interpretation, as the overall mass not in the desired ultrafilter focused on $x$ is equally redistributed to all elements of the filter.

Finally, the unique $L_1/L_2$ partial approximation in $\mathcal{B}$ of Equation (\ref{eq:example-l2-b}) (which is also the barycenter (\ref{eq:example-barycenter-linf-b}) of the partial $L_\infty$ approximations in $\mathcal{B}$) is shown as a red square. Just as the $L_1/L_2$ approximation in $\mathcal{M}$ (green star), it attributes zero mass to $\{x,z\}$, which fails to be supported by any focal element of the original belief function. As a result, they both fall on the border of the tetrahedron of admissible consistent b.f.s focussed on $x$ (highlighted in yellow).

\section{Conclusions} \label{sec:conclusions}

Consistent belief functions represent consistent knowledge bases in the belief logic interpretation of the theory of evidence. Given a belief function inferred from conflicting expert judgments or measurements, the use of a consistent transformation prior to making a decision may be desirable. In this paper we solved the specific instance of the consistent approximation problem associated with measuring distances between belief function using classical $L_p$ norms, in both belief and mass spaces. This is motivated by the fact that consistent belief functions live in a simplicial complex defined in terms of the $L_\infty$ norm. The obtained approximations typically have straightforward interpretations in terms of degrees of belief or mass redistribution. The interpretation of the polytope of all $L_\infty$ solutions is a bit more complex, and worth to be fully investigated in the near future, in the light of the interesting analogy with the polytope of consistent probabilities.

\section*{Appendix}

\textbf{Proof of Theorem \ref{the:l2-m}.} The desired orthogonality condition reads as $\langle \vec{m}_b - \vec{m}_{cs}, \vec{m}_B - \vec{m}_{\{x\}} \rangle = 0$ where $\vec{m}_b - \vec{m}_{cs}$ is given by Equation (\ref{eq:difference-vector-m}), while $\vec{m}_B - \vec{m}_{\{x\}} (C) = 1$ if $C=B$, $=-1$ if $C = \{x\}$, 0 elsewhere. Therefore, using once again the variables $\{ \beta(B) \}$, the condition simplifies as follows:
\begin{equation}\label{eq:condition-l2-n-2}
\langle \vec{m}_b - \vec{m}_{cs}, \vec{m}_B - \vec{m}_{\{x\}} \rangle = \left \{ \begin{array}{ll} \beta(B) - \beta(\{x\}) = 0 & \forall B \supsetneq \{x\}, B \neq \Theta ; \\ -\beta(x) = 0 & B = \Theta. \end{array} \right.
\end{equation}
When using the ($N-1$)-dimensional representation (\ref{eq:dev-m-N-1}) of mass vectors, the orthogonality condition reads instead as:
\begin{equation}\label{eq:condition-l2-n-1}
\langle \vec{m}_b - \vec{m}_{cs}, \vec{m}_B - \vec{m}_{\{x\}} \rangle = \beta(B) - \beta(\{x\}) = 0 \;\;\;\;\; \forall \; B \supsetneq \{x\}.
\end{equation}
In the $N-2$ representation, by (\ref{eq:condition-l2-n-2}) we have that $\beta(B)=0$, i.e., $m_{cs}(B) = m_b(B)$ $\forall B \supseteq \{x\}$, $B \neq \Theta$. By normalization we get $m_{cs}(\Theta) = m_b(\Theta) + m_b(x^c)$: but this is exactly the $L_1$ approximation (\ref{eq:l1-m}).\\
In the $N-1$ representation, the orthogonality condition (\ref{eq:condition-l2-n-1}) reads as
\[
m_{cs}(B) = m_{cs}(x) + m_b(B) - m_b(x)\;\;\; \forall B \supsetneq \{x\}.
\]
By normalizing it we get: $\sum_{\{x\} \subseteq B \subseteq \Theta} m_{cs}(B) = m_{cs}(x) + \sum_{\{x\} \subsetneq B \subseteq \Theta} m_{cs}(B) = 2^{|\Theta|-1} m_{cs}(x) + pl_b(x) - 2^{|\Theta|-1} m_b(x) = 1$, i.e., $m_{cs}(x) = m_b(x) + (1 - pl_b(x))/2^{|\Theta|-1}$, as there are $2^{|\Theta|-1}$ subsets in the ultrafilter containing $x$. By replacing the value of $m_{cs}(x)$ into the first equation we get (\ref{eq:l2-m}).

\textbf{Proof of Lemma \ref{lem:coeff}.} We first note that, by definition of categorical belief function ${b}_A$,
$
\langle \vec{b}_B,\vec{b}_C \rangle = \sum_{D \supseteq B,C; D \neq \Theta} 1 = \sum_{E \subsetneq (B \cup C)^c} 1 = 2^{|(B \cup C)^c|} - 1.
$
Hence:
\begin{equation} \label{eq:quantity}
\begin{array}{l}
\displaystyle
\sum_{B \subseteq A} \langle \vec{b}_B,\vec{b}_C \rangle (-1)^{|B \setminus A|} = \sum_{B \subseteq A} (2^{|(B \cup C)^c|} - 1) (-1)^{|B \setminus A|} \\ = \displaystyle \sum_{B \subseteq A} 2^{|(B \cup C)^c|} (-1)^{|B \setminus A|} - \sum_{B \subseteq A} (-1)^{|B \setminus A|} = \sum_{B \subseteq A} 2^{|(B \cup C)^c|} (-1)^{|B \setminus A|}
\end{array}
\end{equation}
as $\displaystyle \sum_{B \subseteq A} (-1)^{|B \setminus A|} = \sum_{k=0}^{|B \setminus A|} 1^{|A^c|-k} (-1)^k = 0$ by Newton's binomial:
$\displaystyle
\sum_{k = 0}^n \binom{n}{k} p^k q^{n-k} = (p+q)^n.
$
Now, as both $B \supseteq A$ and $C \supseteq A$ the set $B$ can be decomposed into the disjoint sum $B = A + B' + B''$,
where $\emptyset \subseteq B' \subseteq C \setminus A$, $\emptyset \subseteq B'' \subseteq (C \cup A)^c$.\\
Therefore (\ref{eq:quantity}) becomes:
$
\displaystyle
\sum_{\emptyset \subseteq B' \subseteq C \setminus A} \sum_{\emptyset \subseteq B'' \subseteq (C \cup A)^c} 2^{|(A \cup C)|^c - |B''|} (-1)^{|B'| + |B''|} = \\ \displaystyle =
\sum_{\emptyset \subseteq B' \subseteq C \setminus A} (-1)^{|B'|} \sum_{\emptyset \subseteq B'' \subseteq (C \cup A)^c} (-1)^{|B''|} 2^{|(A \cup C)|^c - |B''|}$, where\\
$\displaystyle \sum_{\emptyset \subseteq B'' \subseteq (C \cup A)^c} (-1)^{|B''|} 2^{|(A \cup C)|^c - |B''|} = [2 + (-1)]^{|(A \cup C)|^c} = 1^{|(A \cup C)|^c} = 1$, again by Newton's binomial. The quantity (\ref{eq:quantity}) is therefore equal to
$\displaystyle
\sum_{\emptyset \subseteq B' \subseteq C \setminus A} (-1)^{|B'|},
$
which is nil for $C \setminus A \neq \emptyset$, equal to 1 when $C \subseteq A$.

\textbf{Proof of Theorem \ref{the:global-l1}.} The $L_1$ distance between the partial approximation and $\vec{b}$ can be easily computed as: $\| \vec{b}-\vec{cs}^x_{\mathcal{B},L_1}[b] \|_{L_1} =$
\[
\begin{array}{lll}
& = & \displaystyle \sum_{A \subseteq \Theta} | b(A) - cs^x_{\mathcal{B},L_1}[b](A) | \\ & = & \displaystyle \sum_{A \not\supset \{x\}} | b(A) - 0 | + \sum_{A \supseteq \{x\}} \Big | b(A) - \sum_{B \subseteq A, B \supseteq \{x\}} m_{cs}(B) \Big | \\ & = & \displaystyle \sum_{A \not\supset \{x\}} b(A) + \sum_{A \supseteq \{x\}} \Big | \sum_{B \subseteq A} m_b(B) - \sum_{B \subseteq A, B \supseteq \{x\}} (m_b(B) + m_b(B \setminus \{x\})) \Big | \\ & = & \displaystyle \sum_{A \not\supset \{x\}} b(A) + \sum_{A \supseteq \{x\}} \Big | \sum_{B \subseteq A, B \not \supset \{x\}} m_b(B) - \sum_{B \subseteq A, B \supseteq \{x\}} m_b(B \setminus \{x\}) \Big | \\ & = & \displaystyle \sum_{A \not\supset \{x\}} b(A) + \sum_{A \supseteq \{x\}} \Big | \sum_{C \subseteq A \setminus \{x\}} m_b(C) - \sum_{C \subseteq A \setminus \{x\}} m_b(C) \Big | \\ & = & \displaystyle \sum_{A \not\supset \{x\}} b(A) = \sum_{A \subseteq \{x\}^c} b(A).
\end{array}
\]

\textbf{Proof of Theorem \ref{the:global-l2}.} The $L_2$ distance between the partial approximation and $\vec{b}$ can be easily computed as: $\| \vec{b}-\vec{cs}^x_{\mathcal{B},L_2}[b] \|^2 = $
\[
\begin{array}{lll}
& = & \displaystyle \sum_{A \subseteq \Theta} (b(A) - cs^x_{\mathcal{B},L_2}[b](A))^2 = \sum_{A \subseteq \Theta} \Big ( \sum_{B \subseteq A} m_b(B) - \sum_{B \subseteq A, B \supseteq \{x\}} m_{cs}(B) \Big )^2 \\ & = & \displaystyle \sum_{A \subseteq \Theta} \Big ( \sum_{B \subseteq A} m_b(B) - \sum_{B \subseteq A, B \supseteq \{x\}} m_b(B) - \sum_{B \subseteq A, B \supseteq \{x\}} m_b(B\setminus \{x\}) \Big )^2 \\ & = & \displaystyle \sum_{A \not \supset \{x\}} (b(A))^2 + \sum_{A \supseteq \{x\}} \Big ( \sum_{B \subseteq A, B \not \supset \{x\}} m_b(B) - \sum_{B \subseteq A, B \supseteq \{x\}} m_b(B \setminus \{x\}) \Big )^2 \\ & = & \displaystyle \sum_{A \not \supset \{x\}} (b(A))^2 + \sum_{A \supseteq \{x\}} \Big ( \sum_{C \subseteq A \setminus \{x\}} m_b(C) - \sum_{C \subseteq A \setminus \{x\}} m_b(C) \Big )^2
\end{array}
\]
so that $\displaystyle \| \vec{b}-\vec{cs}^x_{\mathcal{B},L_2}[b] \|^2 = \sum_{A \not \supset \{x\}} (b(A))^2 = \sum_{A \subseteq \{x\}^c} (b(A))^2$.

\bibliographystyle{elsarticle-num}
\bibliography{FSS-simplicial,ai09consistent}

\begin{thebibliography}{10}
\expandafter\ifx\csname url\endcsname\relax
  \def\url#1{\texttt{#1}}\fi
\expandafter\ifx\csname urlprefix\endcsname\relax\def\urlprefix{URL }\fi
\expandafter\ifx\csname href\endcsname\relax
  \def\href#1#2{#2} \def\path#1{#1}\fi

\bibitem{Shafer76}
G.~Shafer, A Mathematical Theory of Evidence, Princeton University Press, 1976.

\bibitem{dempster67multivariate}
A.~Dempster, Upper and lower probabilities induced by a multivariate mapping,
  Annals of Mathematical Statistics 38 (1967) 325--339.

\bibitem{yager87on}
R.~R. Yager, On the dempster-shafer framework and new combination rules,
  Information Sciences 41 (1987) 93--138.

\bibitem{smets81degree}
P.~Smets, The degree of belief in a fuzzy event, Inf. Sciences 25 (1981) 1--19.

\bibitem{DBLP:journals/isci/RamerK93}
A.~Ramer, G.~J. Klir, Measures of discord in the {D}empster-{S}hafer theory.,
  Information Sciences 67~(1-2) (1993) 35--50.

\bibitem{1163941}
W.~Liu, Analyzing the degree of conflict among belief functions, Artif. Intell.
  170~(11) (2006) 909--924.
\newblock \href
  {http://dx.doi.org/http://dx.doi.org/10.1016/j.artint.2006.05.002}
  {\path{doi:http://dx.doi.org/10.1016/j.artint.2006.05.002}}.

\bibitem{hunter06fusion}
A.~Hunter, W.~Liu, Fusion rules for merging uncertain information, Information
  Fusion 7~(1) (2006) 97--134.

\bibitem{lo06mss}
K.~C. Lo, Agreement and stochastic independence of belief functions,
  Mathematical Social Sciences 51(1) (2006) 1--22.

\bibitem{paris08unclog}
J.~B. Paris, D.~Picado-Muino, M.~Rosefield, Information from inconsistent
  knowledge: A probability logic approach, in: Advances in Soft Computing,
  Vol.~46, Springer-Verlag, 2008.

\bibitem{priest89}
G.~Priest, R.~Routley, J.~Norman, Paraconsistent logic: Essays on the
  inconsistent, Philosophia Verlag, 1989.

\bibitem{batens00}
D.~Batens, C.~Mortensen, G.~Priest, Frontiers of paraconsistent logic, in:
  Studies in logic and computation, Vol.~8, Research Studies Press, 2000.

\bibitem{haenni05isipta}
R.~Haenni, Towards a unifying theory of logical and probabilistic reasoning,
  in: Proceedings of ISIPTA'05, 2005.

\bibitem{voorbraak89efficient}
F.~Voorbraak, A computationally efficient approximation of {D}empster-{S}hafer
  theory, International Journal on Man-Machine Studies 30 (1989) 525--536.

\bibitem{cobb03isf}
B.~R. Cobb, P.~P. Shenoy, A comparison of bayesian and belief function
  reasoning, Information Systems Frontiers 5(4) (2003) 345--358.

\bibitem{daniel06on}
M.~Daniel, On transformations of belief functions to probabilities,
  International Journal of Intelligent Systems, special issue on Uncertainty
  Processing.

\bibitem{cuzzolin07smcb}
F.~Cuzzolin, Two new {B}ayesian approximations of belief functions based on
  convex geometry, IEEE Tr. SMC-B 37~(4) (2007) 993--1008.

\bibitem{dubois90}
D.~Dubois, H.~Prade, Consonant approximations of belief functions,
  International Journal of Approximate Reasoning 4 (1990) 419--449.

\bibitem{dubois93possibilityprobability}
D.~Dubois, H.~Prade, S.~Sandri,
  \href{citeseer.ist.psu.edu/dubois93possibilityprobability.html}{On
  possibility/probability transformations} (1993).
\newline\urlprefix\url{citeseer.ist.psu.edu/dubois93possibilityprobability.html}

\bibitem{baroni04ipmu}
P.~Baroni, Extending consonant approximations to capacities, in: IPMU, 2004,
  pp. 1127--1134.

\bibitem{black96examination}
P.~Black, An examination of belief functions and other monotone capacities,
  {PhD} dissertation, Department of Statistics, Carnegie Mellon University,
  pgh. PA 15213 (1996).

\bibitem{cuzzolin08smcc}
F.~Cuzzolin, A geometric approach to the theory of evidence, IEEE Transactions
  on Systems, Man and Cybernetics part C 38~(4) (2008) 522--534.

\bibitem{jousselme10belief}
A.-L. Jousselme, P.~Maupin, On some properties of distances in evidence theory,
  in: Proceedings of BELIEF'10, Brest, France, 2010.

\bibitem{cuzzolin10brest}
F.~Cuzzolin, Geometric conditioning of belief functions, in: Proceedings of
  BELIEF'10, Brest, France, 2010.

\bibitem{shi10distance}
C.~Shi, Y.~Cheng, Q.~Pan, Y.~Lu, A new method to determine evidence distance,
  in: Proceedings of the 2010 International Conference on Computational
  Intelligence and Software Engineering (CiSE), 2010, pp. 1--4.

\bibitem{jiang08new}
W.~Jiang, A.~Zhang, Q.~Yang, A new method to determine evidence discounting
  coefficient, in: Lecture Notes in Computer Science, Vol. 5226/2008, 2008, pp.
  882--887.

\bibitem{khatibi10new}
V.~Khatibi, G.~Montazer, A new evidential distance measure based on belief
  intervals, Scientia Iranica - Transactions D: Computer Science and
  Engineering and Electrical Engineering 17~(2) (2010) 119--132.

\bibitem{diaz06fusion}
J.~Diaz, M.~Rifqi, B.~Bouchon-Meunier, A similarity measure between basic
  belief assignments, in: Proceedings of FUSION'06, 2006.

\bibitem{cuzzolin11isipta-consonant}
F.~Cuzzolin, Lp consonant approximations of belief functions in the mass space,
  in: Proceedings of ISIPTA'11, Innsbruck, Austria, 2011.

\bibitem{cuzzolin11isipta-conditional}
F.~Cuzzolin, Geometric conditional belief functions in the belief space, in:
  Proceedings of ISIPTA'11, Innsbruck, Austria, 2011.

\bibitem{cuzzolin08isaim-consistent}
F.~Cuzzolin, An interpretation of consistent belief functions in terms of
  simplicial complexes, in: Proc. of ISAIM'08, 2008.

\bibitem{cuzzolin10fss}
F.~Cuzzolin, The geometry of consonant belief functions: simplicial complexes
  of necessity measures, Fuzzy Sets and Systems 161~(10) (2010) 1459--1479.

\bibitem{cuzzolin01space}
F.~Cuzzolin, R.~Frezza, Geometric analysis of belief space and conditional
  subspaces, in: Proceedings of the $2^{nd}$ International Symposium on
  Imprecise Probabilities and their Applications (ISIPTA2001), Cornell
  University, Ithaca, NY, 26-29 June 2001.

\bibitem{Saffiotti_abelief-function}
A.~Saffiotti, A belief-function logic, in: Universit Libre de Bruxelles, MIT
  Press, pp. 642--647.

\bibitem{Mates72}
B.~Mates, Elementary Logic, Oxford University Press, 1972.

\bibitem{smets93belief}
P.~Smets, Belief functions : the disjunctive rule of combination and the
  generalized {B}ayesian theorem, International Journal of Approximate
  Reasoning 9 (1993) 1--35.

\bibitem{Novikov_russian}
B.~Dubrovin, S.~Novikov, A.~Fomenko, Sovremennaja geometrija. Metody i
  prilozenija, Nauka, Moscow, 1986.

\bibitem{cuzzolin09isipta-consistent}
F.~Cuzzolin, Consistent approximation of belief functions, in: Proceedings of
  ISIPTA'09, Durham, UK, June 2009.

\bibitem{smets88beliefversus}
P.~Smets, Belief functions versus probability functions, in: S.~L. Bouchon~B.,
  Y.~R. (Eds.), Uncertainty and Intelligent Systems, Springer Verlag, Berlin,
  1988, pp. 17--24.

\bibitem{lewis76}
D.~Lewis, Probabilities of conditionals and conditional probabilities,
  Philosophical Review 85 (1976) 297--–315.

\bibitem{Gardenfors}
P.~Gardenfors, Knowledge in Flux: Modeling the Dynamics of Epistemic States,
  MIT Press, Cambridge, MA, 1988.

\bibitem{perea09amodel}
A.~Perea, A model of minimal probabilistic belief revision, Theory and Decision
  67~(2) (2009) 163--222.

\bibitem{ubf}
P.~Smets, The nature of the unnormalized beliefs encountered in the
  transferable belief model, in: Proceedings of UAI'92, Morgan Kaufmann, San
  Mateo, CA, 1992, pp. 292--29.

\end{thebibliography}







\end{document}